\documentclass[10pt,a4paper,final,logo]{apolloworld}
\usepackage{tikz}
\usepackage{float}
\usepackage{placeins}
\usepackage[numbers,sort&compress]{natbib}
\usepackage[nameinlink,noabbrev]{cleveref}

\makeatletter
\AddToHook{cmd/NAT@sort@cites/after}{\expandafter\Hy@backout\expandafter{\NAT@cite@list}}
\makeatother


\usetikzlibrary{arrows.meta,positioning,fit,backgrounds,calc,shadows,shapes.geometric}
\usepackage{pgfplots}
\pgfplotsset{compat=1.18}
\usepgfplotslibrary{groupplots}

\crefname{equation}{Eq.}{Eqs.}
\crefname{figure}{Fig.}{Figs.}
\crefname{section}{Sec.}{Secs.}
\crefname{appendix}{App.}{Apps.}
\crefname{table}{Tab.}{Tabs.}

\definecolor{awtint}{HTML}{E9F3F0}
\definecolor{awcool}{HTML}{EDF3F5}
\definecolor{awmodelfill}{HTML}{CFE8E1}   % deeper teal tint for learned/model blocks
\definecolor{awdiscfill}{HTML}{FBF3E4}    % warm sand for the discrete side path (contrast)
\definecolor{awoutfill}{HTML}{DCEFF6}     % cool blue for output / read-out heads
\definecolor{awgrpfill}{HTML}{F4F8F7}     % faint panel background for module grouping
\definecolor{awlane}{HTML}{B9C6C2}        % lane strokes in the mini BEV pictogram

\tikzset{
  awshadow/.style = {drop shadow={shadow xshift=0.5pt, shadow yshift=-0.6pt,
                     opacity=0.16, fill=awnight}},
  awbox/.style   = {draw=awblue, line width=0.7pt, rounded corners=3pt, fill=awcool,
                    align=center, inner sep=5pt, font=\scriptsize, awshadow},
  awmodel/.style = {draw=awblue, line width=1.1pt, rounded corners=3pt,
                    fill=awmodelfill, align=center, inner sep=6pt,
                    font=\scriptsize\bfseries, text=awnight, awshadow},
  awio/.style    = {draw=awline, line width=0.6pt, rounded corners=7pt, fill=white,
                    align=center, inner sep=5pt, font=\scriptsize, awshadow},
  awdiscbox/.style = {awbox, fill=awdiscfill, draw=awmuted},
  awhead/.style  = {awbox, fill=awoutfill},
  awgroup/.style = {draw=awblue!45, line width=0.6pt, dashed, rounded corners=6pt,
                    fill=awgrpfill, inner sep=8pt},
  awgrouplbl/.style = {font=\tiny\bfseries, text=awblue, inner sep=1pt},
  awflow/.style  = {-{Latex[length=5pt,width=4.2pt]}, awblue, line width=0.9pt,
                    rounded corners=2pt},
  awfb/.style    = {-{Latex[length=5pt,width=4.2pt]}, awmuted, line width=0.8pt,
                    dashed, rounded corners=2pt},
}

\newcommand{\bevmini}[2]{%
  \begin{tikzpicture}[scale=#1, line join=round, baseline=(bev.center)]
    \node (bev) at (0.5,0.36) {};
    \draw[rounded corners=1.5pt, draw=awline!55, fill=awpaper] (0,0) rectangle (1,0.72);
    \draw[awlane, line width=1.8pt] (0.05,0.22) -- (0.95,0.28);
    \draw[awlane, line width=1.8pt] (0.05,0.50) -- (0.95,0.56);
    \draw[awlane!70, densely dotted, line width=0.4pt] (0.05,0.36) -- (0.95,0.42);
    \fill[awblue] (0.14,0.11) circle (0.9pt);
    \draw[-{Latex[length=2.4pt]}, awblue, line width=0.7pt] (0.15,0.10) -- ++(0.15,0.03);
    \foreach \i/\x/\y in {1/0.34/0.25, 2/0.58/0.53, 3/0.74/0.29, 4/0.46/0.53, 5/0.86/0.55}{%
      \ifnum\i>#2
        \draw[awblue, line width=0.5pt, fill=awtint, rounded corners=0.3pt]
          ($(\x,\y)+(-0.05,-0.032)$) rectangle ($(\x,\y)+(0.05,0.032)$);
      \else
        \fill[apolloworldgreen!75, draw=awblue, line width=0.3pt, rounded corners=0.3pt]
          ($(\x,\y)+(-0.05,-0.032)$) rectangle ($(\x,\y)+(0.05,0.032)$);
      \fi}
  \end{tikzpicture}%
}

\newcommand{\methodname}{Apollo World}

\newcommand{\agenticworld}{Agentic World}
\newcommand{\scenedynamics}{Agent Dynamics}
\newcommand{\scenegen}{Agent Generation}

\newcommand{\generativeworld}{Generative World}

\newcommand{\reporttitle}{\methodname{}: Building Driveable Worlds at Scale}
\newcommand{\reportteam}{World Model Team}
\newcommand{\reportorganization}{Baidu}
\newcommand{\reportdate}{November, 2026}

\title{\reporttitle}
\author{\hyperref[sec:contributors]{\reportteam}\protect\footnotemark\\{\Affilfont \reportorganization}}
\contributornote={A detailed list of contributors and acknowledgments can be found in~\Cref{sec:contributors} of this report.}
\date{\reportdate}

\hypersetup{
  pdftitle    = {\reporttitle},
  pdfauthor   = {\reportteam, \reportorganization},
  pdfsubject  = {An agentic multimodal world model for autonomous driving},
  pdfkeywords = {\methodname{}, world models, autonomous driving, closed-loop simulation, multimodal generation},
  pdfcreator  = {LaTeX with hyperref},
}

\begin{abstract}
Autonomous driving requires experience with rare and unfamiliar situations, as well as interactions that change in response to the vehicle's actions.
Driving logs provide an essential foundation, but each recording captures only one realization of a scene: one route, one set of conditions, and one sequence of interactions.
To model how these interactions unfold under different driving decisions, we introduce \methodname{}, an agentic multimodal world model for autonomous driving.
\methodname{} builds an \emph{\agenticworld{}} in which traffic participants are modeled as interacting AI-driven agents.
Each agent pursues its own goal and adapts its behavior to the ego vehicle and other agents.
These interactions drive world evolution, while \emph{\generativeworld{}} jointly produces the corresponding multi-view camera and LiDAR observations in causal order.
Generation is conditioned on predicted agent states through either layouts or, where reconstructions are available, Gaussian renderings of the reconstructed scene, which provide captured geometry and appearance.
\methodname{} enables autonomy stacks to explore driving worlds beyond recorded experience through closed-loop interaction, from individual scenarios to continuous virtual rides analogous to complete robotaxi trips.
Scenario-level interactions support focused exploration and learning in controlled driving situations; trip-level interactions support training and evaluation through sustained experience as decisions and their consequences unfold over an entire journey.
Please refer to our project page \url{https://apollogo.com/apolloworld} for more results.
\end{abstract}

\begin{document}
\maketitle
\abscontent

\section{Introduction}
\label{sec:introduction}

An autonomous vehicle approaching an unprotected left turn faces several possible futures.
An oncoming car may continue or slow down; a pedestrian may enter the crossing; the vehicle itself may wait or begin the turn.
The vehicle's decision can change how other road users respond and which situations it encounters next.
A recorded drive captures only one outcome.
Developing autonomy stacks requires diverse driving experience beyond existing logs, including interactions under different decisions and varied traffic and visibility conditions.

Apollo Go, Baidu's autonomous ride-hailing service, provides a foundation for learning world dynamics through its real-world operations.
As of October 2026, its fleets had accumulated more than 390 million kilometers of autonomous driving, and its global footprint spanned 30 cities~\citep{baidu-2026-q3-results}.
Each drive contains evidence about road geometry, participant interactions, and their appearance through the vehicle's sensors.
The central question is how to turn this evidence into scalable interactive experience through which the autonomy stack can explore situations beyond the logs, learn from the consequences of its actions, and improve its driving policy.

This requires \emph{driveable worlds}: environments that an autonomy stack can perceive, act within, and revisit under controlled variations.
Traffic must respond plausibly to the ego vehicle.
Cameras and LiDAR must describe a compatible scene.
Observations must arrive in causal order as decisions are made.
The environment must maintain consistent scene state over time to support sustained interaction and learning.

We introduce \methodname{}, an \emph{agentic multimodal world model for autonomous driving} that turns real driving experience into driveable worlds.
It combines interacting AI-driven agents with multimodal sensor generation, connecting driving decisions, agent responses, and sensor observations in a continuous closed loop.

\emph{Traffic agents drive how the world evolves.}
\methodname{} builds an \agenticworld{} in which traffic participants are represented as interacting AI-driven agents.
Each agent pursues its own goal and adapts its behavior to the ego vehicle and other agents.
Agents can be initialized from recorded scenes or generated to create new encounters.
At each step, \agenticworld{} predicts the agents' next states from their histories, road context, and the current ego action.
The updated scene becomes the starting point for subsequent interactions.

\emph{Multimodal generation makes this world perceptible.}
\generativeworld{} jointly synthesizes multi-view camera images and LiDAR measurements from the evolving world and generated history, supplying sensor observations to the autonomy stack.
Camera and LiDAR generation are conditioned on shared scene layouts derived from predicted agent states and road structure, together with semantic conditioning from text prompts.
Cross-sensor attention lets sensors with overlapping fields of view exchange information both across views and across modalities.
Each token carries a sensing-ray embedding that guides attention to relevant features from other sensors.
This helps keep the generated observations consistent while preserving each sensor's native representation.
When a Gaussian reconstruction is available, we can instead condition generation on its renders.
These renders recreate specific real-world scenes for controlled testing, while layout conditioning lets us explore scenarios beyond the reconstructed areas.
The autonomy stack makes its next decision from the generated observations, and the resulting action starts the next round of agent responses and sensor generation.

\begin{figure}[t]
  \centering
  \includegraphics[width=\linewidth]{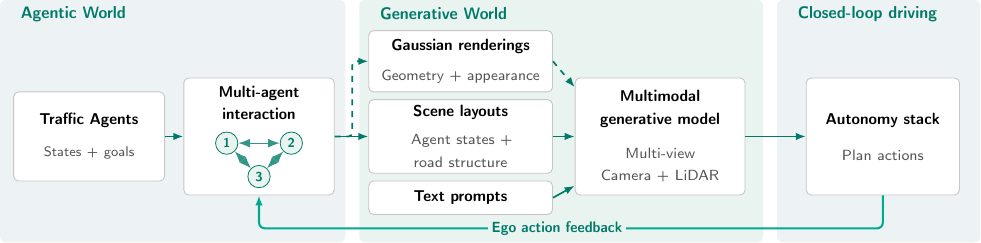}
  \caption{Overview of \methodname{}, an agentic multimodal world model for autonomous driving.
  Traffic participants are modeled as reactive agents initialized from recorded or generated scenes, each pursuing its own goal and responding to ego actions and other agents.
  Predicted agent states and road structure condition joint multi-view camera and LiDAR generation in \generativeworld{} through shared scene layouts.
  Gaussian renderings can also serve as conditions (dashed arrows).
  Text prompts provide further semantic conditioning.
  The autonomy stack acts on generated observations and feeds its actions back to \agenticworld{}.
  This closed loop spans individual scenarios and continuous origin-to-destination virtual rides.}
  \label{fig:apollo-world-overview}
\end{figure}

The resulting worlds support the development, training, and evaluation of autonomous driving systems.
At the scenario level, a recorded event can be replayed and varied to test whether a known issue has been fixed.
This gives a controlled test of an individual interaction.
\methodname{} adds \emph{continuous virtual rides} as a central capability, enabling evaluation of complete driving tasks before deployment in a new operating region, including routes not represented in the driving logs.
Given an origin and a destination, the autonomy stack plans and acts from generated observations in a closed loop.
The route specifies where to go, while the autonomy determines how to drive; local trajectories and maneuvers are not prescribed.
It must handle each situation, carry forward the consequences of earlier decisions, and reach the destination without resets between encounters.

To explain the technical foundations of \methodname{}, we first describe \agenticworld{} and its two components, \scenedynamics{} and \scenegen{} (\Cref{sec:agentic-world}).
We then present \generativeworld{} (\Cref{sec:generative-world}) and Gaussian Scene Reconstruction for grounding (\Cref{sec:gaussian-scene-reconstruction}).
Each section presents the method, learning objective, and experiments.
We next present downstream applications in closed-loop evaluation of safety-critical scenarios and continuous virtual rides (\Cref{sec:applications}).
We discuss related work and its connections to our approach (\Cref{sec:related-work}).
Finally, we summarize our main contributions and outline future~work~(\Cref{sec:conclusion}).

\section{\agenticworld{}}
\label{sec:agentic-world}

We model traffic participants as interacting AI-driven agents with individual states and goals.
\scenedynamics{} models how these agents respond to the ego vehicle and one another, while \scenegen{} creates and completes the traffic population.
Each agent moves through navigable space toward its goal and may slow down, wait for space, or change course as an encounter unfolds.
These interactions drive how the world evolves.

A recorded scene can be revisited with different ego actions while keeping the initial agents and goals fixed.
We can also edit those goals or sample more conservative or aggressive behavior.
\emph{\scenedynamics{}} simulates these encounters with a shared model that jointly predicts the next states of all agents at each step.

\emph{\scenegen{}} creates and completes the surrounding traffic.
It is trained on recorded driving to capture the distribution of traffic layouts given the roads and agents already present.
As the ego travels, new agents must be added beyond the initially observed scene.
For example, vehicles may approach an intersection from side roads or emerge from behind an occluder.
Their motion and visibility should remain consistent with earlier observations, without sudden appearances in road space already observed as empty.

\generativeworld{} (\Cref{sec:generative-world}) uses the predicted agent states and road structure to generate camera and LiDAR observations.
The autonomy stack perceives the scene and plans its next action from these observations.
Its chosen action updates the ego state used by \scenedynamics{} to predict the agents' subsequent behavior.

\subsection{Agent Representation}
\label{sec:agentic-world-agent-representation}

\paragraph{Agent state representation.}
Each agent is represented by a sequence of per-frame states and a goal.
At each frame, the agent state contains two kinds of attributes: continuous attributes, namely position, velocity, heading, acceleration, and size; and discrete attributes, namely category, subtype, and status indicators such as reversing and crossing status.
The state history is the sequence of these states over the observed frames.
The agent's goal specifies a target position and heading, leaving the agent free to wait, yield, follow, or change course in response to nearby traffic.

\vspace{-5pt}

\paragraph{Agent response representation.}
We represent an agent's response as a sequence of discrete motion tokens, each indexing a short motion segment in a vocabulary.
Each segment specifies displacement and heading change relative to its starting pose.
Each agent category (\eg, vehicles, cyclists, or pedestrians) has its own vocabulary of $2048$ entries.
We build each vocabulary from short segments of agent trajectories in driving logs, grouped by motion pattern.
Within each group, k-disks sampling~\citep{trajeglish} repeatedly selects a segment and removes candidates within a trajectory-distance threshold of the selected segment.
The selected segments across all groups form the vocabulary.
Sampling each group separately helps preserve less common motion patterns that sampling all segments together could miss.

To tokenize an agent trajectory, we approximate it one segment at a time.
The matching pose starts at the trajectory's initial position and heading.
At each step, we place each segment from the agent category's vocabulary at this pose.
We compare it with the next target segment using the $L_2$ distance between corresponding bounding-box corners at matching timesteps.
We select a token using this error and kinematic checks, then append it to the sequence.
The segment's final position and heading become the next matching pose.

\begin{samepage}
\subsection{\scenedynamics{}}
\label{sec:agent-dynamics}

\scenedynamics{} models how agents respond to the ego vehicle and one another, conditioned on their state histories and goals (\Cref{sec:agentic-world-agent-representation}), together with map and occupancy context.
It jointly predicts their next states, allowing traffic interactions to evolve as the ego takes action and surrounding agents adjust their behavior.
\end{samepage}

\subsubsection{Model design}
\label{sec:agent-dynamics-model-design}

\begin{figure}[t]
  \centering
  \includegraphics[width=\linewidth]{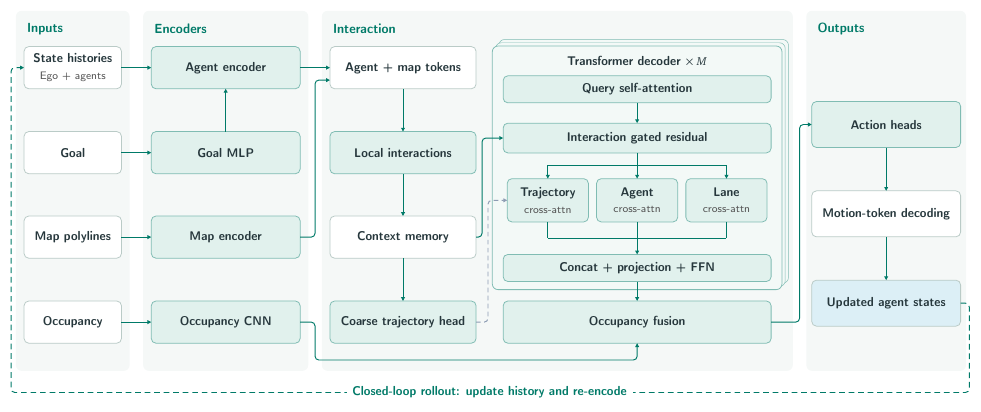}
  \caption{Architecture and closed-loop rollout of \scenedynamics{}.
  State histories (including the ego), goals, and map polylines form tokens whose local interactions produce context memory.
  A coarse trajectory head predicts waypoints from this memory.
  Decoder layers refine agent queries via self-attention, gated interaction features, and parallel local cross-attention to agents, lanes, and predicted waypoints.
  Occupancy features are fused before category-specific action heads predict motion tokens.
  Decoded trajectories advance agent states by one update interval; updated histories are re-encoded at the next step (outer dashed loop).}
  \label{fig:agent-dynamics-architecture}
\end{figure}

\scenedynamics{} uses a shared autoregressive multi-agent Transformer to jointly predict the next states of all agents.
For the $N$ agents whose responses are predicted, let $\mathbf{s}_t$ denote their joint state at time $t$.
Given their state histories $\mathbf{s}_{t-H+1:t}$ over $H$ steps, their goals $\mathbf{g}$, the map and occupancy context $\mathbf{m}$, and the current ego action $\mathbf{a}^{\mathrm{ego}}_t$, the joint prediction is:
\begin{equation}
 P\!\left(\mathbf{s}_{t+1}\mid\mathbf{s}_{t-H+1:t},\mathbf{g},\mathbf{m},\mathbf{a}^{\mathrm{ego}}_t\right).
 \label{eq:agent-dynamics-joint-state-prediction}
\end{equation}
The model first encodes agent state histories, goals, and environmental context, then models interactions among agents and with their surroundings.
A Transformer decoder uses the interaction features to predict motion tokens autoregressively; decoding these tokens yields responses that update the agent states.
These updated states become inputs for the next rollout step.
\Cref{fig:agent-dynamics-architecture} shows the overall architecture and rollout process.

\paragraph{Scene encoding.}
\phantomsection
\label{sec:agent-dynamics-scene-encoding}
Three encoders represent agent histories, map features, and occupancy for interaction and response prediction.

\begin{itemize}
\item \textbf{Agent encoding.}
The agent state encoder maps each agent's state history to a token.
At each timestep, the input comprises the state's continuous and discrete attributes.
Continuous features are projected through a multilayer perceptron (MLP) and combined with learned embeddings of discrete attributes.
Self-attention across timesteps, followed by temporal pooling, then produces an agent token.
The goal is encoded separately and randomly dropped during training to reduce overfitting.

\item \textbf{Map encoding.}
The map encoder processes structured map elements such as lanes, stop lines, and crosswalks.
Each element is represented by a sequence of points with geometric and semantic attributes, such as speed limits, turn directions, lane boundaries, and traffic-light status.
Point-wise features are embedded into a shared feature space and aggregated through attention and convolution.
The resulting features are pooled into one token for each map element.
An auxiliary map loss further supervises the lane tokens.
During training, we randomly mask some lanes and reconstruct them using nearby context.

\item \textbf{Occupancy encoding.}
The occupancy encoder captures the spatial distribution of obstacles and non-drivable regions using an ego-centric grid containing height and categorical information, including hard barriers and construction zones.
A convolutional pyramid progressively reduces the spatial resolution while expanding the feature dimension.
For each agent, the encoder extracts features from the surrounding region of this feature map.
An auxiliary occupancy reconstruction loss further supervises spatial representation learning.
\end{itemize}

\paragraph{Multi-agent interaction.}
Each agent starts with a token summarizing its own state history.
A Transformer interaction module applies a few local attention layers to agent and map tokens.
Each agent gathers neighboring features with weights determined by token features and relative position and heading, using the relative position encoding (RPE) of MTR++~\citep{mtrpp}.
The resulting agent and map features form a context memory for a multi-layer Transformer decoder.
The decoder maintains and refines a query feature for each agent across layers using the context memory.
A coarse trajectory head reads this memory to predict each agent's future trajectory under direct ground-truth (GT) supervision.
The decoder uses reference waypoints from the predicted trajectory, which is detached to prevent decoder losses from backpropagating to the trajectory head.
During training, GT trajectories replace predicted trajectories for a random subset of agents.
The decoder stacks layers that update the per-agent query.
Each layer first applies self-attention among the agent queries.
It then fuses the agent's interaction feature into the query through a gated residual unit, a small MLP whose output passes through a learned gate before being added to the query.
Next, three local cross-attentions run in parallel over neighboring agents, nearby lanes, and the reference waypoints.
Their outputs are concatenated, projected, and added to the query, followed by a feed-forward network with a residual connection.
After the last layer, an MLP fuses each agent's query with the layer's concatenated cross-attention outputs into a decoder output feature $\mathbf{h}_i$.
A separate learned gate then adds occupancy features to $\mathbf{h}_i$, with closer cells receiving larger pooling weights:
\begin{equation}
 \begin{aligned}
 \tilde{\mathbf{o}}_i&=\frac{\sum_{c} w_{ic}\,\mathbf{o}_c}{\sum_{c} w_{ic}+\epsilon_{\mathrm{num}}}, & w_{ic}&=\exp\!\Big(-\tfrac{1}{2}\big\|\mathbf{p}_i-\mathbf{p}_c\big\|^{2}\Big),\\
 \mathbf{h}_i&\;\leftarrow\;\mathbf{h}_i+\sigma\!\big(W_g\tilde{\mathbf{o}}_i+\mathbf{b}_g\big)\odot\operatorname{Norm}\!\big(W_p\tilde{\mathbf{o}}_i\big),
 \end{aligned}
 \label{eq:agent-dynamics-occupancy-gate}
\end{equation}
where the sums run over occupancy cells $c$ in the region surrounding agent $i$, $\mathbf{o}_c$ is the feature of cell $c$, and $\mathbf{p}_i$ and $\mathbf{p}_c$ are agent and cell coordinates normalized by cell size.
$W_g$, $W_p$ and $\mathbf{b}_g$ are learned parameters; $\sigma$ is the sigmoid, $\operatorname{Norm}$ is layer normalization, and $\epsilon_{\mathrm{num}}$ prevents division by zero.
The updated decoder output feature is passed to the action head to predict agent responses.

\paragraph{Agent response prediction and rollout.}
The action head maps each agent's decoder feature to motion-token distributions over future steps.
Each agent category has a separate head over its motion vocabulary.
Default inference selects the highest-probability tokens and decodes them into continuous trajectories.
The runtime advances the agents along these trajectories for one update interval, updating position, heading, velocity, and acceleration in $\mathbf{s}_{t+1}$; size and discrete attributes retain their values from $\mathbf{s}_t$.
The updated state histories are then fed back into the model for the next prediction.
This repeated prediction and update lets agents respond to changes in the ego's action and surrounding traffic.

\paragraph{Response validation and fallback.}
At each update, the runtime validates every predicted agent response before applying it, checking for collisions, occupancy crossings, drivable-area departures, and stalling, among other issues.
If a check fails, a lightweight spatiotemporal planner takes over and jointly optimizes path and speed over a short horizon.
Its objective balances agreement with the model prediction against safe distances, road compliance, progress, and smoothness.
An optimized planner response replaces the model prediction only if it passes the same checks.
The model continues inference during takeover, and once its prediction passes the checks again, \scenedynamics{} regains control.
The fallback can be disabled to measure the model's responses on their own, as reported in \Cref{sec:agent-dynamics-evaluation}.

\subsubsection{Learning agent responses}
\label{sec:agent-dynamics-learning}

Learning to match a recorded continuation is only part of the task.
During simulation, agents encounter state histories produced by their own earlier responses.
Small prediction errors can accumulate, leading to interactions that depart from the log.
We therefore train in three stages.
Pretraining learns behavior from recorded motion; closed-loop supervised finetuning exposes the model to its own generated histories; and reinforcement finetuning adds explicit objectives for avoiding collisions and occupancy crossings.

\paragraph{Behavior cloning pretraining.}
We pretrain the model on GT histories with a token-prediction loss $\mathcal L_{\mathrm{token}}$, a trajectory-matching loss $\mathcal L_{\mathrm{traj}}$, and auxiliary losses $\mathcal L_{\mathrm{aux}}$ for map reconstruction, occupancy reconstruction, and the coarse trajectory head, supervised by the corresponding GT trajectories:
\begin{equation}
 \mathcal L_{\mathrm{pre}}
 =\mathcal L_{\mathrm{token}}+\mathcal L_{\mathrm{traj}}
 +\mathcal L_{\mathrm{aux}}.
 \label{eq:agent-dynamics-pretraining-loss}
\end{equation}
Let $\theta$ be the model parameters, $N$ the number of agents whose responses are predicted, $a_i$ the motion token predicted for agent $i$, and $a_i^{\ast}$ the target token assigned to agent $i$'s GT motion (\Cref{sec:agentic-world-agent-representation}).
At each prediction step, the token prediction loss $\mathcal L_{\mathrm{token}}$ trains the model to predict these targets:
\begin{equation}
 \mathcal{L}_{\mathrm{token}} = -\sum_{i=1}^{N} \log P_\theta\!\left(a_i=a_i^{\ast} \,\middle|\, \mathbf{s}_{t-H+1:t},\mathbf{g},\mathbf{m},\mathbf{a}^{\mathrm{ego}}_t\right).
 \label{eq:agent-dynamics-motion-token-loss}
\end{equation}

For $\mathcal L_{\mathrm{traj}}$, we decode each agent's predicted motion token into a trajectory and compare it with the GT trajectory:
\begin{equation}
 \mathcal L_{\mathrm{traj}}
 =\frac{1}{N}\sum_{i=1}^{N}
 \left\|\hat{\mathbf q}_i-\mathbf q_i\right\|_2,
 \label{eq:agent-dynamics-trajectory-loss}
\end{equation}
where $\hat{\mathbf q}_i$ and $\mathbf q_i$ are the predicted and GT trajectories of agent $i$.

\paragraph{Closed-loop supervised finetuning.}
During pretraining, the model uses GT state histories as inputs.
During rollout, it updates these histories with its own predicted states, so prediction errors can accumulate over multiple steps.
Closed-loop finetuning feeds predicted states back as training inputs~\citep{self-forcing,catk}.
We retain the pretraining objective and add a multi-step rollout loss $\mathcal L_{\mathrm{roll}}$ with weight $\lambda_{\mathrm{roll}}$:
\begin{equation}
 \mathcal L_{\mathrm{SFT}}
 =\mathcal L_{\mathrm{pre}}+\lambda_{\mathrm{roll}}\,\mathcal L_{\mathrm{roll}}.
 \label{eq:agent-dynamics-supervised-fine-tuning-loss}
\end{equation}
The rollout loss $\mathcal L_{\mathrm{roll}}$ applies the trajectory-matching criterion in \Cref{eq:agent-dynamics-trajectory-loss} over multiple autoregressive steps, comparing the predicted rollout trajectories with GT motion.

At each rollout step, we select the highest-probability motion token for each agent and decode it to obtain the next state.
We feed these predicted states into the next step, stopping gradients through this input path.
To let trajectory errors update the token probabilities, we use a straight-through estimator.
Let $p_j$ be the probability of token $j$, $\mathbf{e}_j$ its continuous motion, and $\mathcal K$ the top-$32$ token indices:
\begin{equation}
 \begin{gathered}
  k=\arg\max_j p_j,\quad \mathbf{e}=\mathbf{e}_k,\qquad
  \tilde{\mathbf{e}}=\frac{\sum_{j\in\mathcal K}p_j \mathbf{e}_j}{\sum_{j\in\mathcal K}p_j},\\
  \mathbf{e}_{\mathrm{ST}}=\tilde{\mathbf{e}}+\operatorname{sg}[\mathbf{e}-\tilde{\mathbf{e}}],
 \end{gathered}
 \label{eq:agent-dynamics-straight-through-estimator}
\end{equation}
where $\operatorname{sg}$ stops gradients.
The straight-through motion $\mathbf{e}_{\mathrm{ST}}$ is used to compute $\mathcal L_{\mathrm{roll}}$: its forward value equals the selected motion $\mathbf{e}$, while its gradients flow through the probability-weighted motion $\tilde{\mathbf{e}}$.
We also add noise to the agent state histories during training to improve robustness to prediction errors during rollout.

\begin{table}[t]
  \centering
  \caption{Reward terms and weights used in reinforcement finetuning.
  All terms are penalties.}
  \label{tab:agent-dynamics-reward-weights}
  \begin{tabularx}{\linewidth}{>{\raggedright\arraybackslash}p{.20\linewidth}Y>{\raggedleft\arraybackslash}p{.10\linewidth}}
    \toprule
    \textbf{Reward term} & \textbf{Penalty criterion} & \textbf{Weight} \\
    \midrule
    Collision & Bounding-box overlap with another agent & $-10.0$ \\
    Proximity & Clearance to another agent below a safety margin & $-2.0$ \\
    Occupancy crossing & Agent footprint crossing a hard boundary & $-10.0$ \\
    Lateral deviation & Lateral offset from the GT trajectory & $-1.0$ \\
    Trajectory deviation & Clipped ADE relative to the GT trajectory & $-0.2$ \\
    Travel efficiency & Speed below the GT speed by a margin & $-0.15$ \\
    \bottomrule
  \end{tabularx}
\end{table}

\paragraph{Reinforcement finetuning.}
The supervised objectives in pretraining and closed-loop finetuning do not directly penalize undesirable outcomes such as collisions and occupancy crossings, so reinforcement finetuning optimizes reward terms that penalize them on the model's own rollouts.
We update the Transformer decoder and the action heads, while freezing the encoders and the local interaction module.
A category-aware critic estimates values, and generalized advantage estimation~\citep{gae} uses $\gamma=0.99$ and $\lambda=0.95$.
Clipped PPO~\citep{ppo} with $\epsilon=0.2$ updates the policy relative to a frozen reference decoder.
The objective includes value regression, entropy regularization, and a reference-policy KL penalty with $\beta_{\mathrm{KL}}=0.3$.
Action imitation and rollout-state imitation remain active with weights $0.5$ and $0.1$ to maintain consistency with recorded behavior.
Rewards penalize collisions, insufficient clearance, occupancy crossings, deviation from recorded motion, and inefficient progress, as detailed in \Cref{tab:agent-dynamics-reward-weights}.

\subsubsection{Controlling agent behavior}
\label{sec:agent-dynamics-behavior-control}
We control agent behavior through intent guidance and behavior sampling.
A goal specifies a target position and heading, leaving arrival time and speed unspecified.
Each agent can adjust its path and speed in response to surrounding traffic.
By default, the model selects the most probable motion token.
To generate varied behavior, we instead sample from high-probability tokens, adjusting their probabilities using motion aggressiveness and interaction danger levels.
Motion aggressiveness reflects speed, acceleration, and turning, while interaction danger reflects the minimum distance or time to collision between the agent and the ego.
These levels steer sampling toward conservative or aggressive behavior, allowing the same scene to be simulated with more accommodating or more assertive agents and thereby adjusting the interaction pressure on the ego.

\subsubsection{Model evaluation}
\label{sec:agent-dynamics-evaluation}

We evaluate whether \scenedynamics{} produces accurate and realistic motion, responds to ego actions, and supports behavior control.
Using recorded driving data, we report trajectory accuracy and assess behavioral realism.
We then test interactions in simulation, where the ego is controlled by the onboard autonomy stack used on Apollo Go robotaxis.
The agents are reactive and controlled by \scenedynamics{}.
We compare this setup with non-reactive replay to assess conflict resolution when the ego departs from the recording.
We also vary sampling aggressiveness to assess how behavior control changes the interaction pressure on the ego.

\paragraph{Datasets.}
Training uses driving data collected across our global operations, including $\sim712$ million nominal frames and $\sim504$ million mined interaction-tagged frames.
The data cover a range of driving scenarios, including turns, merges, roundabouts, and U-turns.
We use these tags to sample interactions and less common behaviors more frequently during training.
Trajectories are smoothed to correct tracking artifacts and improve kinematic consistency before being used to construct the motion vocabulary and generate token labels.
The smoothing step uses nonlinear optimization with bounds on acceleration and curvature rate.
Evaluation uses a nominal set of $\sim1.25$ million frames and a safety-critical set of $\sim0.49$ million frames containing recorded traffic conflicts.

\begin{table}[t]
  \centering
  \caption{Offline performance across training stages on the nominal evaluation set.
  SFT denotes closed-loop supervised finetuning and RFT denotes reinforcement finetuning.}
  \label{tab:agent-dynamics-training-stage-results}
  \begin{tabularx}{\linewidth}{>{\raggedright\arraybackslash}p{.43\linewidth}RRR}
    \toprule
    \textbf{Metric} & \textbf{Pretraining} & \textbf{SFT} & \textbf{RFT} \\
    \midrule
    ADE (m) $\downarrow$ & $0.79$ & $0.74$ & $0.78$ \\
    FDE (m) $\downarrow$ & $1.58$ & $1.47$ & $1.53$ \\
    Collision rate (\%) $\downarrow$ & $1.60$ & $1.51$ & $0.60$ \\
    Ego-collision rate (\%) $\downarrow$ & $0.34$ & $0.32$ & $0.07$ \\
    Goal-achieved rate (\%) $\uparrow$ & $77.3$ & $81.4$ & $83.4$ \\
    Occupancy-crossing rate (\%) $\downarrow$ & $0.32$ & $0.30$ & $0.26$ \\
    \bottomrule
  \end{tabularx}
\end{table}

\begin{table}[t]
  \centering
  \caption{Closed-loop collision outcomes in simulation with non-reactive replay and reactive agent responses.}
  \label{tab:agent-dynamics-closed-loop-collisions}
  \begin{tabularx}{\linewidth}{>{\raggedright\arraybackslash}p{.34\linewidth}RRR}
    \toprule
    \raisebox{-.5\baselineskip}{\textbf{Evaluation set}}
      & \textbf{Replay\linebreak\mbox{collision pass rate}}
      & \textbf{Reactive\linebreak\mbox{collision pass rate}}
      & \raisebox{-.5\baselineskip}{\textbf{Repair rate}} \\
    \midrule
    Nominal evaluation set & $92.32\%$ & $98.94\%$ & $86.54\%$ \\
    Safety-critical evaluation set & $0\%$ & $93.68\%$ & $93.68\%$ \\
    \bottomrule
  \end{tabularx}
\end{table}

\begin{table}[t]
  \centering
  \caption{Ego-collision rates by aggressiveness on a mined subset with dense interactions.}
  \label{tab:agent-dynamics-aggressiveness-results}
  \begin{tabularx}{\linewidth}{>{\raggedright\arraybackslash}p{.43\linewidth}RRR}
    \toprule
    \textbf{Metric} & \textbf{Conservative} & \textbf{Default (top-1)} & \textbf{Aggressive} \\
    \midrule
    Ego-collision rate & $2.1\%$ & $3.8\%$ & $15.6\%$ \\
    \bottomrule
  \end{tabularx}
\end{table}

\begin{figure}[t]
  \centering
  \includegraphics[width=\linewidth]{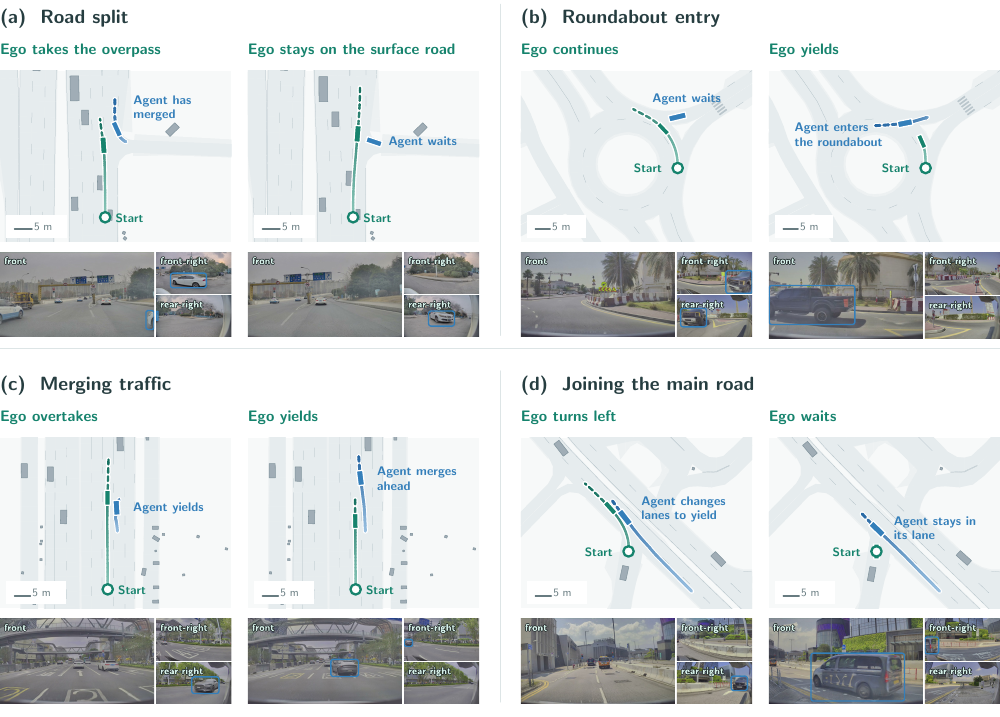}
  \vspace{-20pt}
  \caption{Illustrative agent responses under alternative ego actions from a shared starting world state.
  Each pair uses the same map view and scale.
  Lines trace ego (green) and key-agent (blue) motion through each rollout.
  Lines are solid up to the displayed frame and dashed afterward.
  \generativeworld{} (\Cref{sec:generative-world}) generates the camera views shown for each state (front at left, front-right at upper right, and rear-right at lower right).}
  \label{fig:agent-dynamics-ego-action-responses}
  \vspace{-5pt}
\end{figure}

\paragraph{Evaluation protocol.}
Trajectory accuracy is measured by average displacement error (ADE) and final displacement error (FDE) over four-second horizons.
Rollout metrics include collision, occupancy-crossing, and goal-achievement.
Collision pass rate is the fraction of scenes completed without collision; repair rate is the fraction of replay-collision cases made collision-free by reactive agents.

\paragraph{Agent response quality.}
\Cref{tab:agent-dynamics-training-stage-results} shows that closed-loop supervised finetuning improves response accuracy on the nominal evaluation set, reducing four-second ADE from $0.79$\,m to $0.74$\,m.
Reinforcement finetuning lowers collision rate from $1.51\%$ to $0.60\%$ and raises goal achievement from $81.4\%$ to $83.4\%$, while ADE increases to $0.78$\,m.
On our evaluation set, the Waymo Open Sim Agents Challenge realism meta-metric~\citep{wosac} is $0.7502$, measuring distributional similarity to recorded driving.
\Cref{fig:agent-dynamics-ego-action-responses} illustrates how agents merge, wait, yield, or change lane under alternative ego actions.
The figure shows their states and motion alongside synchronized generated camera observations.

\paragraph{Reactive responses and behavior control.}
Reactive agents raise the nominal collision pass rate from $92.32\%$ to $98.94\%$ and repair $86.54\%$ and $93.68\%$ of replay-collision cases on the nominal and safety-critical sets, respectively (\Cref{tab:agent-dynamics-closed-loop-collisions}).
In offline evaluation on a mined subset with dense interactions, ego-collision rate changes from $3.8\%$ under default top-1 selection to $2.1\%$ under the most conservative setting and $15.6\%$ under the most aggressive setting (\Cref{tab:agent-dynamics-aggressiveness-results}).
Agents with conservative behavior are more likely to yield, while those with aggressive behavior put pressure on the ego.

\FloatBarrier
\subsection{\scenegen{}}
\label{sec:agent-generation}

Reactive simulation requires a plausible population of agents, including those not yet fully observed.
As the ego moves, vehicles may approach from side roads, emerge from occlusion, or enter the sensing range.
\scenegen{} generates this population conditioned on the map and any \textsc{known} agents.
Each generated agent is described by a sequence of current and past states, together with a goal.
The generated scene provides the initial conditions for \scenedynamics{} and the scene structure for sensor generation.

\subsubsection{Model design}
\label{sec:agent-generation-model-design}

\scenegen{} uses the agent representation in \Cref{sec:agentic-world-agent-representation}.
The input contains map context and a fixed number of agent slots of three types.
\textsc{known} slots contain context agents, and \textsc{generated} slots contain agents to synthesize.
Unfilled slots are marked \textsc{padding} and masked, as are slots containing invalid agents or agents outside the perception range.
The model fills active \textsc{generated} slots conditioned on the map, ego state, and any \textsc{known} agents.
The \textsc{generated} slot count sets the output agent count.
An optional binary lane mask guides generation toward selected regions, such as along the ego's route or near occluded areas.

\begin{figure}[t]
  \centering
  \includegraphics[width=\linewidth]{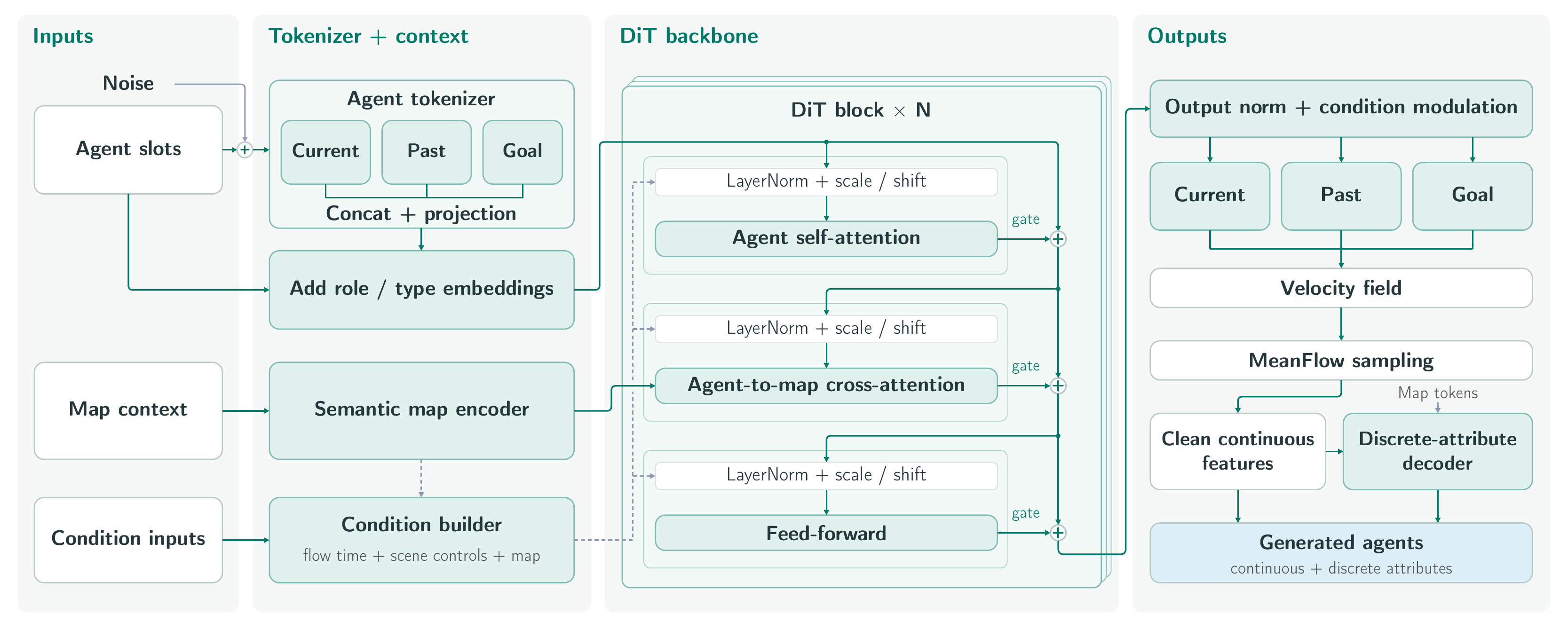}
  \caption{Architecture of \scenegen{}.
  Agent slots are tokenized from continuous current-state (\emph{Current}), past-state (\emph{Past}), and goal (\emph{Goal}) features at flow time $t$.
  Before tokenization, \textsc{generated} features start as Gaussian noise for sampling or GT mixed with noise for training; \textsc{known} and \textsc{padding} stay clean.
  Each token receives \emph{role} (ego or other agent) and \emph{slot-type} (\textsc{known}, \textsc{generated}, or \textsc{padding}) embeddings.
  A condition builder combines flow time, slot-based agent counts, an optional generation region, and a map summary to modulate every DiT block through adaLN-Zero scale, shift, and gate parameters.
  Sampling integrates the DiT's MeanFlow (average) velocities into clean features.
  These are \textsc{generated} agents' continuous attributes; a separate map-conditioned decoder predicts discrete attributes from them.}
  \label{fig:agent-generation-architecture}
\end{figure}

The model generates continuous and discrete attributes separately (\Cref{fig:agent-generation-architecture}).
Let $x$ collect generated agents' continuous current-state, past-state, and goal features, and $y$ their discrete attributes: category, subtype, and status indicators.
Condition $c$ comprises the map, ego state, features of \textsc{known} surrounding agents, and an optional generation-region mask.
Their joint conditional distribution is factorized as:
\begin{equation}
 p(x,y\mid c)=p_{\mathrm{cont}}(x\mid c)\,p_{\mathrm{disc}}(y\mid x,c).
 \label{eq:agent-generation-factorization}
\end{equation}
We model $p_{\mathrm{cont}}$ with MeanFlow~\citep{meanflow}, using a conditional Diffusion Transformer (DiT) backbone.
Starting from Gaussian noise, the model generates $x$ in a few sampling updates.
Flow time indexes progress from noise to the generated features, and the sampling interval sets each update's span.
The discrete-attribute decoder models $p_{\mathrm{disc}}$, predicting $y$ from the generated $x$ and context $c$.

\paragraph{Network architecture.}
The tokenizer receives a fixed-length sequence of agent slots: \textsc{known} slots retain clean features, \textsc{generated} slots carry noisy features, and \textsc{padding} slots are masked.
Each slot is tokenized by separately projecting its continuous current-state, past-state, and goal features through three input branches and fusing their outputs.
The current-state features are position, heading, velocity, size, and acceleration.
Each of six past frames contains displacement and heading change relative to the current pose.
The goal specifies target position and heading.
We normalize continuous attributes such as position, velocity, and size by subtracting their training-set means and dividing by their standard deviations.
Headings use sine--cosine pairs.
A role embedding distinguishes the ego from other agents, and a slot-type embedding marks each slot as \textsc{known}, \textsc{generated}, or \textsc{padding}.

A map encoder tokenizes polylines into map tokens; the optional region mask marks selected lanes.
Agent tokens pass through DiT blocks, each applying agent self-attention, map cross-attention, and a feed-forward network.
adaLN-Zero~\citep{dit} conditions these blocks on flow time, sampling interval, agent counts, and summaries of the map and \textsc{known} agents.
For a specified generation region, its lane-token is added to this condition.

Three heads map normalized DiT outputs to current-state, past-state, and goal feature updates at each sampling step.
After sampling, inverse normalization restores attributes in their original units.
A separate discrete-attribute decoder predicts \textsc{generated} agents' discrete attributes from generated features and map context.
The decoder learns to infer each agent's category from its generated size and motion.

\paragraph{Sampling.}
Flow time $t$ runs from noisy initialization ($t=1$) to completed samples ($t=0$).
Let $z_t$ denote all agent slots' normalized continuous features at flow time $t$.
Sampling transforms \textsc{generated} slots from Gaussian noise into normalized $x$.
For target flow time $r<t$ and sampling interval $t-r$, the DiT's output heads jointly predict $u_\theta(z_t,t,r,c)$, the average change in $z_t$ per unit of flow time (average velocity) under condition $c$, where $\theta$ denotes the model parameters.
Following MeanFlow~\citep{meanflow}, we update the generated components of $z_t$ as:
\begin{equation}
 z_r=z_t-(t-r)u_\theta(z_t,t,r,c).
 \label{eq:agent-generation-sampling}
\end{equation}
We update until $t=0$.
Throughout sampling, \textsc{known} agents remain fixed, and \textsc{padding} slots are masked.
We then restore the generated features to original units, constrain the continuous attributes, keep \textsc{generated} agents within the perception range, and predict their discrete attributes with the discrete-attribute decoder.

\subsubsection{Learning agent generation}
\label{sec:agent-generation-learning}

Training considers valid surrounding agents within the perception range.
We sample tasks that generate either all of these agents or a selected subset, such as agents that are occluded, newly detected, or entering from intersections or merging lanes.
Selected agents are \textsc{generated}; the remaining valid agents are \textsc{known}.

\paragraph{Continuous-attribute supervision.}
We use the notation of \Cref{eq:agent-generation-sampling}, with $z_0$ denoting the normalized GT continuous features across all agent slots.
During training, we construct $z_t$ by mixing the \textsc{generated} agents' features with Gaussian noise $\epsilon$~\citep{flow-matching}:
\begin{equation}
 z_t=(1-t)z_0+t\epsilon,\qquad v=\epsilon-z_0.
 \label{eq:agent-generation-flow-interpolation}
\end{equation}
\textsc{known} agents use $\epsilon=z_0$, so their features remain unchanged for every $t$.
Here $v=dz_t/dt$ is the change per unit of flow time.
We sample a target flow time $r$ with $0\le r\le t$ for each training input.
The model's prediction $u_\theta(z_t,t,r,c)$ is supervised using the MeanFlow target~\citep{meanflow}:
\begin{equation}
 u_{\mathrm{target}}
 =\operatorname{sg}\!\left[v-(t-r)\frac{d}{dt}u_\theta(z_t,t,r,c)\right],
 \qquad \frac{d}{dt}u_\theta(z_t,t,r,c)
 =\frac{\partial u_\theta}{\partial z_t}v+\frac{\partial u_\theta}{\partial t}.
 \label{eq:agent-generation-mean-flow-target}
\end{equation}
A Jacobian-vector product evaluates the derivative with $r$ and $c$ fixed; $\operatorname{sg}$ stops gradients through the target.

For scene $s$'s \textsc{generated} agents $\mathcal G_s$, with predictions $u_{\theta,i}$ and targets $u_{\mathrm{target},i}$ from \Cref{eq:agent-generation-mean-flow-target}, the loss is:
\begin{equation}
 \mathcal L_{\mathrm{cont}}
 =\frac{1}{|\mathcal G_s|}
   \sum_{i\in\mathcal G_s}
   \left\|u_{\theta,i}-u_{\mathrm{target},i}\right\|_{W}^2
   +0.1\mathcal L_{\mathrm{known}}.
 \label{eq:agent-generation-continuous-attribute-loss}
\end{equation}
Here $\|\mathbf{d}\|_{W}^2=\mathbf{d}^{\top}W\mathbf{d}$, where diagonal $W$ upweights each agent's current-position components.
The auxiliary loss $\mathcal L_{\mathrm{known}}$ averages the same per-agent loss over \textsc{known} agents to preserve their supplied features.
On some training examples, we independently omit map conditioning, \textsc{known}-agent conditioning, and the generation-region mask, enabling generation with or without these signals and classifier-free guidance~\citep{cfg}.

\paragraph{Discrete-attribute supervision.}
The discrete-attribute decoder uses cross-entropy losses for category, subtype, and status indicators of \textsc{generated} agents, with continuous-feature inputs detached from the flow backbone.
Training gradually replaces noisy GT features with model-generated features, exposing the decoder to feature errors at inference.

\subsubsection{Model evaluation}
\label{sec:agent-generation-evaluation}

Generated populations need not reproduce a particular recording.
We evaluate whether their agents fit the road, have consistent physical states and motion, cover diverse behaviors, and resemble recorded traffic.

\paragraph{Datasets.}
\label{sec:agent-generation-datasets}
Training uses the same global driving corpus as \scenedynamics{} (\Cref{sec:agent-dynamics-evaluation}).
The main evaluation uses a completion set of $296{,}696$ scenes containing $7.45$ million valid agents.
We use GT scenes as a reference for evaluating the physical consistency and traffic statistics of generated scenes.
We also evaluate road consistency for creation and completion on a separate validation set.

\paragraph{Evaluation metrics.}
We evaluate trajectory regularity, kinematic consistency, and history safety to measure physical consistency.
We also compare behavior in recorded and generated traffic and check how generated vehicles fit the road.
\begin{itemize}[itemsep=1pt]
  \item \textbf{Trajectory regularity.} Fraction of agents passing four checks. No step between successive positions exceeds the $99$th percentile of recorded step lengths.
  The path turns by at most $120^\circ$ between steps, overall motion is within $90^\circ$ of the agent's heading, and its current center lies no more than $5$\,m outside the road corridor.
  \item \textbf{Kinematic consistency.} Fraction of agents within the limits on speed, longitudinal acceleration, and lateral acceleration.
  Each limit is the $99$th percentile of that quantity in recorded data.
  For each agent, the speed in its current state and the average speed inferred from its past positions must differ by at most $8$\,m/s.
  \item \textbf{History safety.} At each past time step, we compare each agent's bounding box with those of other agents in the scene.
  We report the fraction of pairs with a gap of at least $0.5$\,m where the two boxes are closest.
  \item \textbf{Behavior-mode coverage.} For each road layout (e.g., a straight road or an intersection), we compute the fraction of recorded behavior modes (stationary, moving straight, or turning) that also appear in generated traffic.
  We obtain the overall coverage $C$ by averaging these fractions with equal weight for each road layout.
  \item \textbf{Behavior-diversity score.} For each road layout, we compute the Jensen--Shannon divergence (JSD) between the recorded and generated distributions of behavior modes.
  Similarity across layouts is $S=\operatorname{mean}(1-\mathrm{JSD})$.
  The diversity score $2CS/(C+S)$ is the harmonic mean of mode coverage $C$ and distributional similarity $S$.
  \item \textbf{Road consistency.} \textit{Off-road rate} is the fraction of dynamic generated vehicles whose centers are over $4$\,m from the nearest lane centerline.
  Within that distance, the \textit{wrong-way rate} is the share facing against lane travel.
  \textit{Heading error} is the mean heading deviation from the nearest lane segment, ignoring travel direction.
  \textit{Collision rate} is the share of generated vehicles whose oriented boxes intersect another valid agent's box.
\end{itemize}

\paragraph{Scene consistency and diversity.}
On the evaluation set, $99.40\%$ of generated agents pass the trajectory-regularity checks, compared with $98.70\%$ of GT agents (\Cref{tab:agent-generation-quality}).
The kinematic-consistency pass rate is $95.84\%$ versus $96.33\%$ in GT, and the history-safety pair pass rate is $99.42\%$ versus $99.72\%$ in GT.
The model covers $99.40\%$ of recorded behavior modes on average across layouts.
With relative behavior frequencies included, the diversity score is $92.00\%$.

\begin{table}[t]
  \centering
  \caption{Agent-generation quality compared against GT scenes.
  All values are percentages; higher is better.}
  \label{tab:agent-generation-quality}
  \vspace{-5pt}
  \begin{tabularx}{\linewidth}{>{\raggedright\arraybackslash}p{.50\linewidth}RR}
    \toprule
    \textbf{Metric} & \textbf{GT} & \textbf{\scenegen{}} \\
    \midrule
    Trajectory-regularity pass & $98.70$ & $99.40$ \\
    Kinematic-consistency pass & $96.33$ & $95.84$ \\
    History-safety pass & $99.72$ & $99.42$ \\
    Behavior-mode coverage & $100.00$ & $99.40$ \\
    Behavior-diversity score & $100.00$ & $92.00$ \\
    \bottomrule
  \end{tabularx}
\end{table}

\begin{table}[t]
  \centering
  \caption{Road consistency of \textsc{generated} vehicles under creation and completion.
  Lower is better.
  Off-road and wrong-way rates are evaluated on dynamic vehicles.}
  \label{tab:agent-generation-road-consistency}
  \vspace{-5pt}
  \begin{tabular*}{\linewidth}{@{\extracolsep{\fill}\hspace{\tabcolsep}}lrrrr}
    \toprule
    \textbf{Setting} & \textbf{Off-road~(\%)} & \textbf{Wrong-way~(\%)} & \textbf{Heading error~($^\circ$)} & \textbf{Collision~(\%)} \\
    \midrule
    Creation & $1.32$ & $3.38$ & $2.37$ & $2.36$ \\
    Completion & $1.20$ & $3.87$ & $1.70$ & $2.30$ \\
    \bottomrule
  \end{tabular*}
\end{table}

\begin{figure}[t]
  \centering
  \vspace{5pt}
  \includegraphics[width=\linewidth]{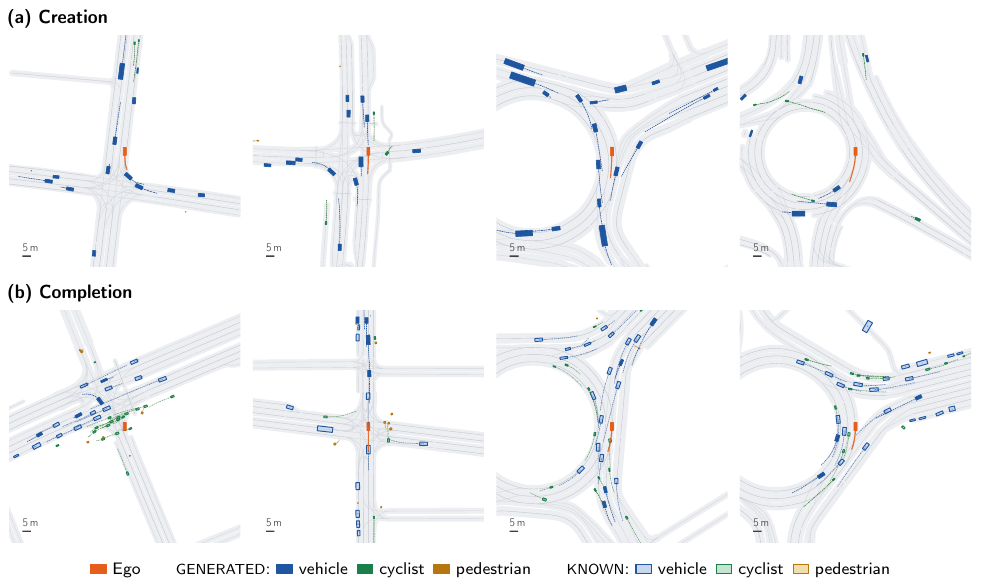}
  \vspace{-15pt}
  \caption{Generated scenes across diverse road layouts.
  (a)~Creation: generating all agents given the map and ego state.
  (b)~Completion: adding agents conditioned on the map, ego state, and \textsc{known} agents.
  Color encodes category and shade encodes role; box orientation gives heading, and fading lines trace recent motion.}
  \label{fig:agent-generation-road-layouts}
\end{figure}

\paragraph{Agents in road context.}
\Cref{tab:agent-generation-road-consistency} reports off-road rates of $1.32\%$ for creation and $1.20\%$ for completion, with collision rates of $2.36\%$ and $2.30\%$, respectively.
Wrong-way rates are $3.38\%$ for creation and $3.87\%$ for completion.
The corresponding mean heading errors relative to the nearest lane are $2.37^\circ$ and $1.70^\circ$.
\Cref{fig:agent-generation-road-layouts} shows generated scenes across a range of road layouts, with agents placed mostly within the drivable area and oriented along feasible paths.

\section{\generativeworld{}}
\label{sec:generative-world}

\generativeworld{} converts scene states predicted by \agenticworld{} into synchronized camera and LiDAR observations for the autonomy stack.
Consistency across views, modalities, and time is essential: sensor discrepancies can change the autonomy stack's decisions, other agents' responses, and the world's evolution.
We use a single Diffusion Transformer (DiT) to jointly generate pinhole images, fisheye images, and LiDAR measurements.
All sensors share one backbone, while sensor-specific projections preserve modality-specific information.
Rendered road and agent layouts or Gaussian reconstructions ground the two modes (\Cref{fig:generative-architecture}): layout-conditioned and Gaussian-conditioned generation, both with the same sensor output interface.

\subsection{Model design}
\label{sec:generative-model-design}

\subsubsection{Sensor-native representations}
\label{sec:generative-sensor-representations}

The sensor rig comprises six pinhole cameras, four fisheye cameras, and four LiDARs.
Each generated stream preserves its sensor's native sampling structure without being reprojected onto a common image plane.

\begin{itemize}
\item \textbf{Pinhole cameras.}
Pinhole cameras cover the surroundings of the ego vehicle with perspective images.
Each stream is an RGB video in its own image plane, defined by its intrinsics and mounting pose on the vehicle.

\item \textbf{Fisheye cameras.}
Fisheye cameras cover the near field around the vehicle with wide-angle images.
Their strong radial distortion is kept: each stream remains an RGB video under the calibrated fisheye projection.

\item \textbf{LiDAR.}
Each LiDAR uses a native range image: rows follow beam inclinations, columns follow azimuth, and cells hold range, return validity, and intensity.
Calibration converts these images directly to point clouds.
\end{itemize}

\begin{figure}[t]
  \centering
  \includegraphics[width=\linewidth]{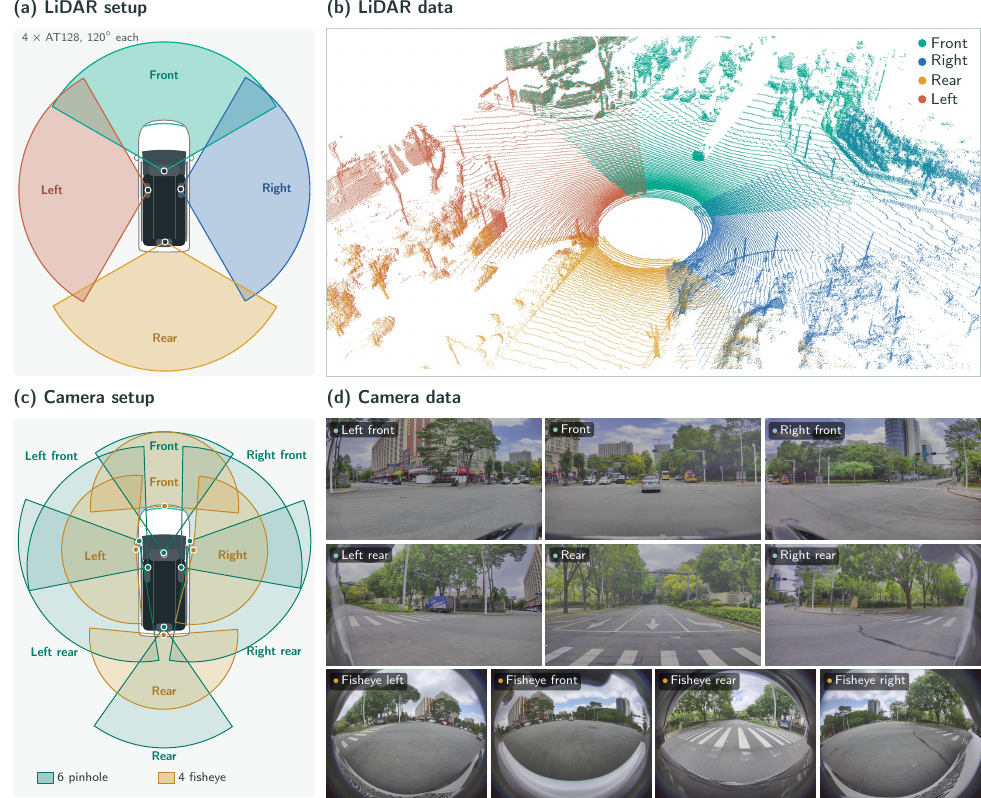}
  \caption{Sensor rig and a capture from a held-out scene.
  (a)~Mounting positions and horizontal FOVs of four LiDARs (120$^\circ$ each, about 30$^\circ$ overlap between neighbors).
  (b)~LiDAR returns from one frame merged in vehicle coordinates and colored by sensor; angular sampling is coarser in the near field (0.5--7.2\,m).
  (c)~Mounting and horizontal FOVs of six pinhole cameras: about 70$^\circ$ front/rear, 100$^\circ$ forward-looking sides, and 120$^\circ$ rearward-looking sides; four fisheye cameras: about 190$^\circ$.
  (d)~All ten camera images from the same frame.}
  \label{fig:generative-sensor-rig}
\end{figure}

All sensors are synchronized at the logging rate of 10\,Hz.
\Cref{fig:generative-sensor-rig} shows our sensor setup.

\subsubsection{Conditions}
\label{sec:generative-conditioning}

Rendered scene conditions guide the generation of sensor observations.
Layout renders encode road structure and agent positions, while Gaussian renders provide geometry and appearance from a reconstructed scene.
The model also uses reference observations, text prompts, sensing rays, and observation history as conditions.

\begin{figure}[t]
  \centering
  \includegraphics[width=\linewidth]{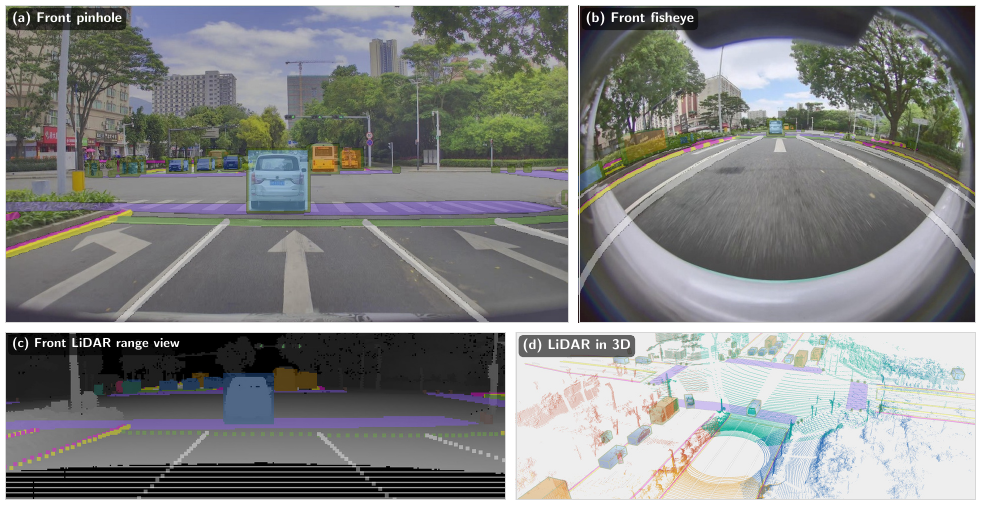}
  \caption{Layout renders for one frame of a held-out scene.
  (a)~Front pinhole and (b)~front fisheye images with the control video overlaid.
  (c)~Front LiDAR range-view control overlaid on the LiDAR range image (near bright, far dark).
  Returns are sparse in the lower rows due to near-field downsampling in our LiDARs.
  (d)~A 3D visualization of the layout control on the four-LiDAR point cloud.}
  \label{fig:generative-layout-renders}
\end{figure}

\begin{itemize}
\item \textbf{Layout renders (Layout-conditioned mode).}
Lane lines, road boundaries, crosswalks, and the 3D boxes of agents are rendered into a control video for every sensor using its calibrated projection; for LiDAR, they are rendered on the range image.
\Cref{fig:generative-layout-renders} shows the three sensor types on one frame; the 3D view in panel~(d) is for illustration only and is not a model input.
Editing an agent trajectory therefore changes all sensors consistently, and editing the ego trajectory changes both the rendered controls and the sensing rays below.
Each control video is encoded by a video VAE~\citep{wan}, passed through a zero-initialized patch embedding, and added to the target tokens at the same positions.

\item \textbf{Gaussian renders (Gaussian-conditioned mode).}
When a Gaussian reconstruction (\Cref{sec:gaussian-scene-reconstruction}) is available, we place the agents' Gaussian assets according to the agent layout and render them together with the background Gaussian asset.
A ControlNet-style branch~\citep{controlnet} encodes the render and adds zero-initialized residual conditioning features to the backbone, which is trained together with the branch (\Cref{sec:generative-end-to-end-fine-tuning}).

\item \textbf{Reference observations.}
When supplied, captured observations at the start of the sequence anchor the scene's appearance and geometry.
Their clean latents replace the noisy latents at the reference positions, which are marked by a mask channel and excluded from the loss.

\item \textbf{Text.}
Prompts describe the time of day, weather, and environment.
They are encoded by a frozen text encoder~\citep{cosmos-reason1} and read through cross-attention in every block.

\item \textbf{Sensing rays.}
Each camera pixel and LiDAR range-image cell has a sensing ray with an origin and direction.
All rays are expressed in the vehicle coordinate system at the first frame.
They therefore combine the intrinsics, extrinsics, and ego pose for each frame, and encode both where each sensor looks and how the vehicle moves.
The ray features enter the tokens before cross-sensor attention (\Cref{sec:generative-cross-sensor-interaction}).

\item \textbf{Observation history.}
We generate sensor observations in causal order, conditioned on earlier observations.
During rollout and inference, the model caches and reuses the keys and values of the generated history.
\end{itemize}

\begin{figure}[t]
  \centering
  \includegraphics[width=\linewidth]{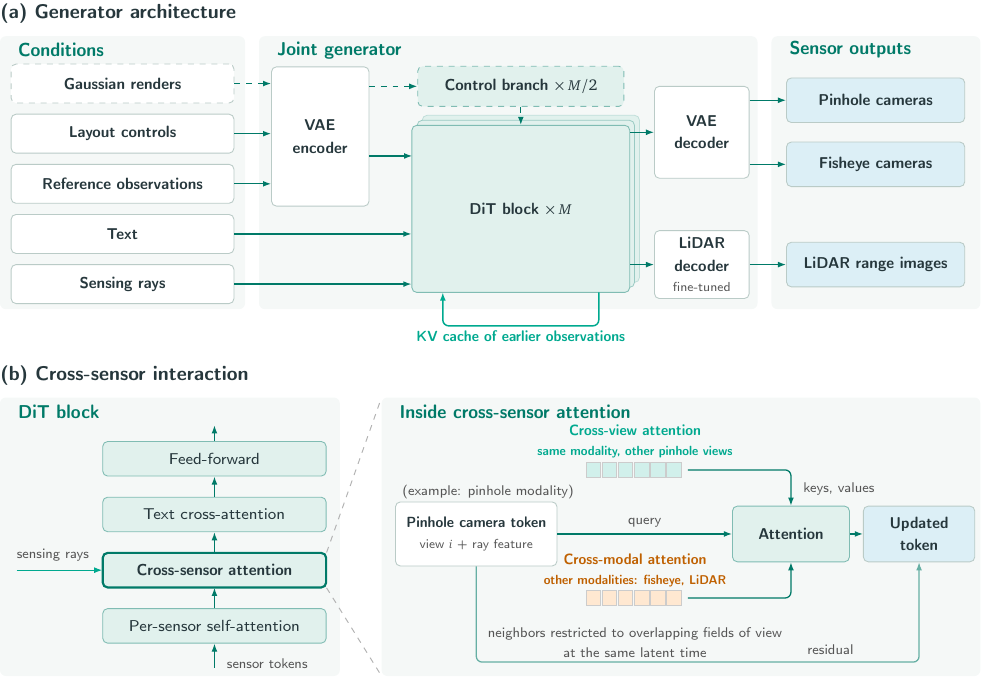}
  \caption{Overview of \generativeworld{} architecture.
  Dashed elements are used only in the Gaussian-conditioned mode.
  (a)~Rendered conditions and reference observations are encoded by the VAE encoder, while text and sensing rays enter the DiT blocks directly.
  Gaussian renders enter through a control branch that mirrors every other DiT block.
  A LiDAR decoder fine-tuned on range maps decodes the LiDAR, and cached keys and values of earlier observations serve as observation history.
  (b)~In selected blocks, a token of the sensor being generated, augmented with its sensing ray, queries the overlapping tokens of other views of the same modality (cross-view attention, green) and of the other modalities (cross-modal attention, orange) at the same latent time; the result is added back as a residual.
  The illustrated query comes from a pinhole-camera token; fisheye and LiDAR tokens are generated in the same way.}
  \label{fig:generative-architecture}
\end{figure}

\subsubsection{Architecture}
\label{sec:generative-architecture}

In \Cref{fig:generative-architecture}a, the VAE encoder~\cite{wan} maps all sensors and rendered conditions into a shared latent space.
Cameras enter as RGB videos; LiDAR range, validity, and intensity form three-channel videos that pass through a mapping layer before encoding.
Cameras use the original VAE decoder~\cite{wan}.
LiDAR decoding combines a VAE decoder and refinement network fine-tuned on range maps to recover range, return validity, and intensity.

A single transformer backbone processes the joint sensor latents after patch embedding.
Each block applies self-attention within each sensor, cross-attention to the text condition, and a feed-forward layer.
In selected blocks, cross-sensor exchange at synchronized times is inserted between the self-attention and the text cross-attention (\Cref{sec:generative-cross-sensor-interaction}).
Because all streams share the latent layout, all sensors are generated in one joint forward pass.

Sensor-specific adaptations give the model capacity for differences in projection and signal statistics without duplicating the backbone.
Camera streams are modulated by view identity; fisheye streams share a type-specific low-rank adaptation, applied after the cross-sensor exchange; the LiDAR path adds a modality embedding, feature modulation, a geometry projection, and low-rank residual adapters.

The model incorporates the conditions described in \Cref{sec:generative-conditioning} as follows.
Rendered controls and the reference mask enter at the token embedding stage.
Text conditions enter through cross-attention, and sensing-ray features are added to the token features used for cross-sensor attention.
Gaussian renders are encoded by the control branch, which injects their features into the backbone.

\subsubsection{Geometry-aware cross-sensor interaction}
\label{sec:generative-cross-sensor-interaction}

Sensors observe the same scene from different viewpoints and with different projections, so the same object may appear at different positions in the sensor observations or be visible to only some sensors.
We construct a sensor graph from sensor placement and fields of view, connecting pairs whose views are expected to overlap.
Each sensor attends to its graph neighbors at the same latent time (\Cref{fig:generative-architecture}b).
We apply Fourier feature maps to each sensing ray's origin and unit direction in the fixed reference frame, concatenate them, and project to the token dimension with a linear layer.
Attention combines ray and sensor features to learn which evidence from another sensor is relevant.
Let $u_{s,t,p}$ be spatial token $p$ of sensor $s$ at latent time $t$, $r_{s,t,p}$ its ray feature, and $\mathcal N(s)$ the set of neighboring sensors.
With $a_{s,t,p}=\operatorname{LayerNorm}(u_{s,t,p}+r_{s,t,p})$, the cross-sensor attention update is:
\begin{equation}
 u'_{s,t,p}=u_{s,t,p}+W_o\,
 \operatorname{Attn}\!\left(W_q a_{s,t,p},
 \{W_k a_{j,t,q}:j\in\mathcal N(s),\,q\in\mathcal T_j\},
 \{W_v a_{j,t,q}:j\in\mathcal N(s),\,q\in\mathcal T_j\}\right),
 \label{eq:generative-cross-sensor-attention}
\end{equation}
where $\mathcal T_j$ contains all spatial tokens of neighbor $j$.
The graph decides which sensors can be read, while the sensing rays and learned attention decide where to attend within them.

\subsection{Model training}
\label{sec:generative-training}

\begin{figure}[t]
  \centering
  \includegraphics[width=\linewidth]{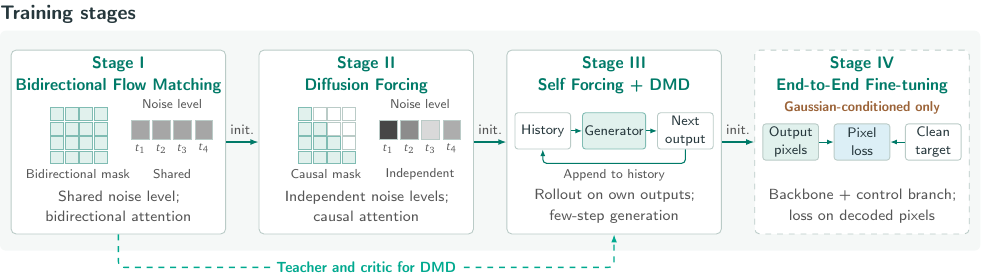}
  \caption{Training stages of \generativeworld{}.
  Stages I--III turn a bidirectional prior into a few-step causal generator, with teacher and critic initialized from Stage I.
  Gaussian-conditioned generation follows I--III on degraded--clean pairs, then Stage IV in pixel space.
  Attention grids index query and key time steps (rows and columns); gray tiles mark noise levels, shared in Stage I and independently sampled over time in Stage II.}
  \label{fig:generative-training-stages}
\end{figure}

The pretrained VAE encoder~\citep{wan}, camera decoder, and text encoder remain frozen.
We fine-tune the LiDAR decoding module on range maps.
We train the multimodal DiT in three stages (\Cref{fig:generative-training-stages}).
Stage I learns a bidirectional joint prior over all sensors with flow matching.
Stage II pairs Diffusion Forcing with causal attention to adapt this prior for autoregressive generation.
Stage III trains on self-generated history and uses distribution matching distillation for few-step rollouts with less error accumulation.
Stage IV uses pixel-space fine-tuning to improve decoded image quality in the Gaussian-conditioned setting, where we add a ControlNet-style branch (\Cref{sec:generative-end-to-end-fine-tuning}).
We follow the training recipe of OmniDreams~\citep{omnidreams} for Stages I--III.
Stages I and II randomly use zero, one, or two initial latent frames as clean references.
Stage III initializes each rollout with one clean latent reference frame.
Throughout Stages I--III, we exclude the reference positions from the loss.

\subsubsection{Stage I: Bidirectional Flow Matching}
\label{sec:generative-bidirectional-flow-matching}

Stage I learns a joint prior over all sensors with flow matching~\citep{flow-matching}.
For a clean latent sequence $z$ and Gaussian noise $\epsilon$, the noisy input is $z^\sigma=(1-\sigma)z+\sigma\epsilon$, where $\sigma\in[0,1]$ denotes the noise level shared across all sensors and frames.
The model predicts the velocity with bidirectional attention over the whole sequence.
To balance the contributions of the three modalities, we average errors within each modality before combining the losses:
\begin{equation}
 \mathcal L_{\mathrm{FM}}=\mathbb E\Bigl[\sum_{g\in\{\mathcal P,\mathcal F,\mathcal D\}}\lambda_g
 \bigl\|v_\theta(z^\sigma,\sigma;c)-(\epsilon-z)\bigr\|_g^2\Bigr],
 \label{eq:generative-flow-matching-loss}
\end{equation}
where $\mathcal P$, $\mathcal F$, and $\mathcal D$ denote pinhole, fisheye, and LiDAR, $\|\cdot\|_g^2$ averages the squared error over all latent elements of modality $g$, $\lambda_g$ are modality weights, and $c$ collects the conditions of \Cref{sec:generative-conditioning}.
The shared backbone, cross-sensor layers, and sensor-specific adaptations are trained jointly.
The resulting bidirectional model is used to initialize the Stage-II generator and serves as the teacher for Stage III.

\subsubsection{Stage II: Causal Diffusion Forcing}
\label{sec:generative-causal-diffusion-forcing}
In closed-loop simulation, each new observation must be generated before the autonomy stack can determine its next action.
Stage II adapts the Stage-I model with Diffusion Forcing~\citep{diffusion-forcing} and causal attention: each prediction attends only to the current and earlier observations.
Noise levels are sampled independently over time and shared across synchronized sensor streams.
With $z_k$ denoting the sensor latents at frame $k$ and $\boldsymbol\sigma$ the per-frame noise levels, the objective keeps the modality-balanced form of \Cref{eq:generative-flow-matching-loss}:
\begin{equation}
 \mathcal L_{\mathrm{DF}}=\mathbb E\Bigl[\sum_{g\in\{\mathcal P,\mathcal F,\mathcal D\}}\lambda_g
 \bigl\|v_\theta(z^{\boldsymbol\sigma}_{\le k},\boldsymbol\sigma_{\le k};c_{\le k})-(\epsilon_k-z_k)\bigr\|_g^2\Bigr],
 \label{eq:generative-diffusion-forcing-loss}
\end{equation}
The expectation also averages uniformly over frame indices $k$.
Stage II thus adapts the bidirectional Stage-I model into a causal generator.
However, inference drift remains because training still uses noisy GT history rather than the model's own outputs.
Stage III mitigates this drift by training on self-generated history.

\subsubsection{Stage III: Self Forcing with DMD}
\label{sec:generative-self-forcing-distillation}

We apply Self Forcing~\citep{self-forcing} by rolling out the Stage-II generator during training: new observations are generated with a few denoising steps conditioned on the model's earlier outputs, with gradients kept at a randomly selected denoising step.
Distribution matching distillation (DMD)~\citep{dmd} supervises the rollout.
A frozen real-score model (the teacher) and a trainable fake-score model (the critic), both initialized from Stage I, evaluate the generated trajectory $\widetilde z$ after re-noising as $y_\sigma=(1-\sigma)\widetilde z+\sigma\epsilon$, where $\epsilon\sim\mathcal N(0,I)$ is independently sampled.
We train the generator with the following DMD surrogate loss:
\begin{equation}
 \mathcal L_{\mathrm{DMD}}=\frac12\mathbb E
 \left\|\widetilde z-
 \operatorname{sg}\!\left(\widetilde z-
 \frac{\widehat z_{\mathrm{fake}}(y_\sigma)-\widehat z_{\mathrm{real}}(y_\sigma)}
 {\operatorname{mean}|\widetilde z-\widehat z_{\mathrm{real}}(y_\sigma)|}
 \right)\right\|_2^2,
 \label{eq:generative-distillation-loss}
\end{equation}
where $\widehat z_{\mathrm{real}}$ and $\widehat z_{\mathrm{fake}}$ are the clean estimates of the two score models.
The stop-gradient operator $\operatorname{sg}$ treats the target as constant during backpropagation.
Both the denominator and the squared-error norm average over all latent elements of each sample.
Both score models process the complete multi-sensor sequence bidirectionally, providing DMD supervision over the sequence as a whole.
The fake-score model is updated by flow matching on detached rollouts to track the current generator, alternating with the generator updates.

At inference, the generator uses four denoising steps for each prediction and reuses the KV cache of earlier observations within an attention window, so long sequences can be continued without recomputing the history.

\subsubsection{Stage IV: End-to-End Fine-tuning}
\label{sec:generative-end-to-end-fine-tuning}

For Gaussian-conditioned generation, the backbone starts from the layout-conditioned weights and is trained jointly with the control branch through Stages I--III with the same objectives on degraded--clean pairs (\Cref{sec:generative-evaluation}), yielding bidirectional, causal, and few-step Gaussian-conditioned generators that correct the renders.

Latent-space objectives do not directly penalize blur in the decoded images.
To sharpen Gaussian-conditioned outputs, Stage IV fine-tunes the backbone and the control branch end-to-end in pixel space, through the few-step generator and the VAE decoder, on the same pairs.
The generated camera videos $\widehat x$ are decoded by the VAE and compared with the clean targets $x$ using a reconstruction, a perceptual, and a style loss, following~\citep{difix3d}:
\begin{equation}
 \mathcal L_{\mathrm{E2E}}=\|\widehat x-x\|_1
 +\lambda_{\mathrm{LPIPS}}\,\mathrm{LPIPS}(\widehat x,x)
 +\lambda_{\mathrm{Gram}}\sum_l\bigl\|G_l(\widehat x)-G_l(x)\bigr\|_2^2,
 \label{eq:generative-end-to-end-fine-tuning-loss}
\end{equation}
where LPIPS~\citep{lpips-eval} measures perceptual distance and $G_l$ is the Gram matrix of layer-$l$ features of a pretrained image network.
The Gram term matches feature correlations between the generated and target images, which favors realistic texture in the corrected regions (\Cref{tab:generative-gaussian-conditioned-results}).

\subsection{Model evaluation}
\label{sec:generative-evaluation}

We evaluate the layout-conditioned mode on multi-sensor generation and the Gaussian-conditioned mode on camera correction and harmonization.
The evaluation covers pinhole images, fisheye images, and LiDAR outputs.
We also assess text-based weather and lighting control, as well as cross-view and temporal consistency.

\paragraph{Datasets.}
The layout-conditioned mode is trained on 55k driving scenes, each with 30 seconds of data recorded at 10\,Hz ($\sim$16.5 million frames), each with 6 pinhole cameras + 4 fisheye cameras + 4 LiDARs.
The scenes are carefully curated to cover diverse weather, lighting, environments, and traffic conditions.
We evaluate multi-sensor generation on a held-out set of 200 diverse scenes, with no train–evaluation overlap or temporal continuity.

The Gaussian-conditioned mode is trained on aligned degraded--clean camera pairs for two tasks: a \emph{Fixer} removes Gaussian artifacts, and a \emph{Harmonizer} adapts inserted assets to the existing scene.
The \emph{Fixer} training data cover 700 reconstructed scenes of 10 seconds each.
\emph{Fixer pairs} follow Difix3D+~\citep{difix3d}, using captured views as targets and renders from under-optimized reconstructions or perturbed camera poses as inputs.
The \emph{Harmonizer} training data contain 600 re-insertion scenes, each 10 seconds long.
\emph{Harmonizer pairs} use captured views as targets.
Inputs are constructed by re-inserting an object with a changed appearance.
The object is either cut out and re-inserted with randomized color and lighting in image space after shadow removal, or reconstructed as a digital asset~\citep{sam3d} and rendered back into the 3DGS background.

\paragraph{Evaluation protocol.}
For the layout-conditioned mode, we report Fr\'echet Inception Distance (FID)~\citep{fid-eval} and Fr\'echet Video Distance (FVD)~\citep{fvd-eval}.
The initial GT frame is excluded from scoring.
LiDAR evaluation uses returns between 3 and 70\,m.
We report Fr\'echet distance in Utonia feature space (FUD)~\citep{utonia} without intensity input, Chamfer distance between generated and GT point clouds of the same frame, return-validity F1 on the native beam grid, and Wasserstein-1 distance (W1) between valid-return intensity distributions, averaged over range bins.

For the Gaussian-conditioned mode, we report PSNR, SSIM, and LPIPS between generated camera videos and the corresponding GT to measure reconstruction accuracy and perceptual similarity.
FID and FVD assess image and video realism, respectively.
The evaluation covers scenes with rendering artifacts, re-inserted assets, or both.
Scores are computed for each scene and then averaged equally across scenes.

\begin{table}[t]
  \centering
  \caption{Layout-conditioned multimodal generation with the bidirectional teacher model (Stage I, 35 steps) and the causal student model (Stage III, 4 steps).
  Pinhole and fisheye scores are averaged over six and four cameras, respectively.
  Bold marks the better point estimate in each column.}
  \label{tab:generative-layout-conditioned-results}
  \begin{tabular*}{\linewidth}{@{\extracolsep{\fill}\hspace{\tabcolsep}}lrrrrrrrr}
    \toprule
    & \multicolumn{2}{c}{\textbf{Pinhole}} & \multicolumn{2}{c}{\textbf{Fisheye}} & \multicolumn{4}{c}{\textbf{LiDAR}} \\
    \cmidrule(lr){2-3}\cmidrule(lr){4-5}\cmidrule(lr){6-9}
    Model & FID $\downarrow$ & FVD $\downarrow$ & FID $\downarrow$ & FVD $\downarrow$ & FUD $\downarrow$ & \shortstack[r]{Chamfer\\(m) $\downarrow$} & \shortstack[r]{Return\\F1 $\uparrow$} & \shortstack[r]{Intensity\\W1 $\downarrow$} \\
    \midrule
    Bidirectional & \textbf{11.0} & \textbf{38.8} & \textbf{11.4} & 48.7 & 66.74 & \textbf{0.275} & \textbf{0.951} & 4.51 \\
    Causal & 11.2 & 45.0 & 12.4 & \textbf{43.4} & \textbf{56.96} & 0.333 & 0.947 & \textbf{4.46} \\
    \bottomrule
  \end{tabular*}
\end{table}
\begin{table}[t]
  \centering
  \caption{Comparison with OmniDreams, with metrics reported only for the front pinhole camera.
  We finetune its released single-view checkpoint on our data and distill a causal student using the released configuration.
  Our models jointly generate ten camera views and four LiDAR streams.
  Best scores are in bold.}
  \label{tab:generative-front-camera-comparison}
  \begin{tabular*}{\linewidth}{@{\extracolsep{\fill}\hspace{\tabcolsep}}lrrrr}
    \toprule
    & \multicolumn{2}{c}{\textbf{Bidirectional teacher}} & \multicolumn{2}{c}{\textbf{Causal student}} \\
    \cmidrule(lr){2-3}\cmidrule(lr){4-5}
    Model & FID $\downarrow$ & FVD $\downarrow$ & FID $\downarrow$ & FVD $\downarrow$ \\
    \midrule
    OmniDreams~\citep{omnidreams} & 12.83 & 55.64 & 13.53 & 61.45 \\
    Ours & \textbf{9.01} & \textbf{38.75} & \textbf{9.47} & \textbf{43.24} \\
    \bottomrule
  \end{tabular*}
\end{table}

\paragraph{Layout-conditioned generation.}
\Cref{tab:generative-layout-conditioned-results} compares the bidirectional teacher with the causal generator on the layout-conditioned evaluation set.
Each scene is a 10-second clip conditioned on the first GT frame and text prompts.
This comparison examines how distillation from a 35-step teacher into a four-step causal generator affects multimodal generation quality.
The causal generator achieves lower FVD than the bidirectional teacher for fisheye cameras (43.4 versus 48.7) but higher FVD for pinhole cameras (45.0 versus 38.8).
LiDAR FUD decreases from 66.74 to 56.96 and intensity W1 from 4.51 to 4.46, while Chamfer distance increases from 0.275 to 0.333\,m.
Return F1 decreases from 0.951 to 0.947.
We also compare with OmniDreams in \Cref{tab:generative-front-camera-comparison}.
Since only a single-view checkpoint is publicly released~\citep{omnidreams-code}, the comparison is limited to front-camera FID and FVD.
We finetune this teacher checkpoint on our data and distill a causal student following the released configuration.
Our FID/FVD scores are 9.01/38.75 and 9.47/43.24 for the bidirectional teacher and four-step causal student, respectively, compared with 12.83/55.64 and 13.53/61.45 for OmniDreams.
Both our models achieve lower scores while jointly generating ten camera views and four LiDAR streams.

\begin{figure}[t]
  \centering
  \includegraphics[width=\linewidth]{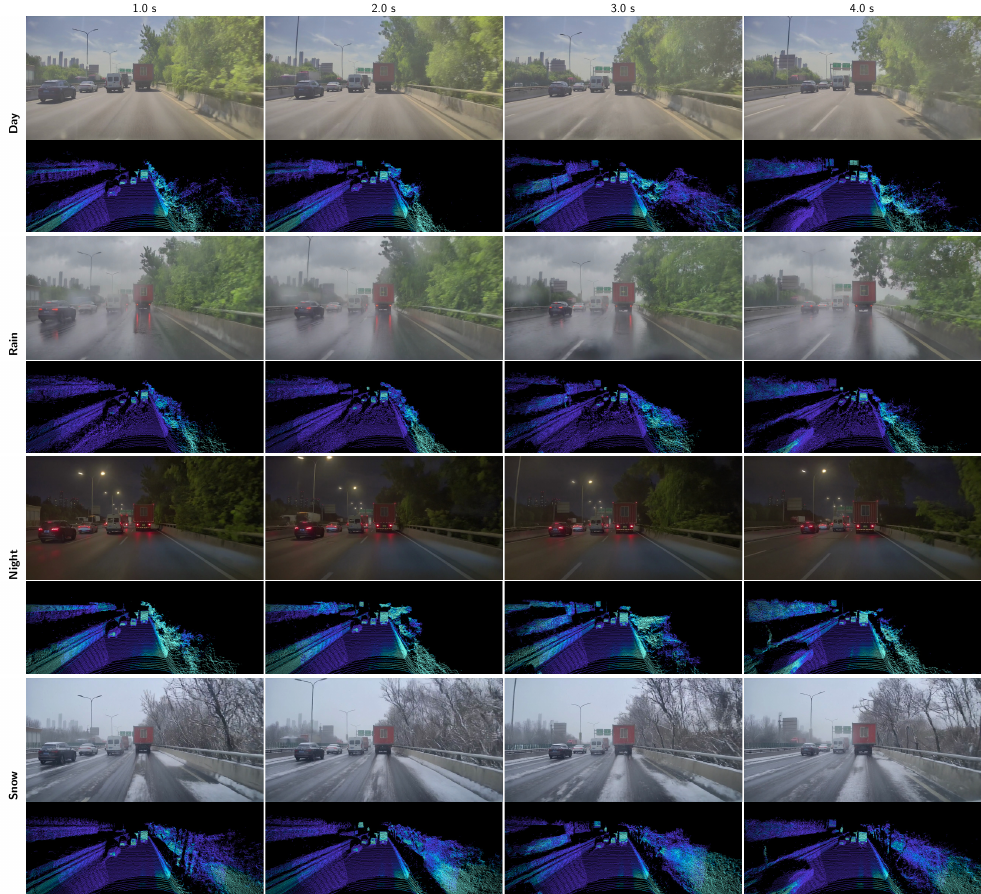}
  \caption{Weather and lighting control.
  All rollouts share the layout controls and the seed.
  For each prompt, the upper row shows the generated front camera and the lower row the generated LiDAR at 1, 2, 3, and 4\,s.}
  \label{fig:generative-weather-lighting-control}
\end{figure}

\paragraph{Weather and lighting control.}
\Cref{fig:generative-weather-lighting-control} keeps the scene and its layout controls fixed and changes the text prompt between daytime, heavy rain, clear night, and snow.
The pinhole cameras receive one initial reference frame consistent with the prompt, while the fisheye cameras and LiDAR receive none, so their appearance follows the prompt and the evidence exchanged with the cameras.
In rain, the road becomes reflective and distant structures fade into spray; at night, street lamps, headlights, and tail lights become the dominant light sources; in snow, the shoulders, guardrails, and branches turn white while the lanes stay wet with tire tracks.
The LiDAR follows the prompt as well: in rain, the returns from the wet road surface around the vehicle become sparse and dark, as standing water reflects the beams away from the sensor.
Across all four prompts, the truck ahead, the surrounding vehicles, and the gantry signs appear at the same positions at every timestamp, because they are fixed by the shared layout controls.
Text therefore changes the appearance of the scene without changing its content.

\begin{figure}[t]
  \centering
  \includegraphics[width=\linewidth]{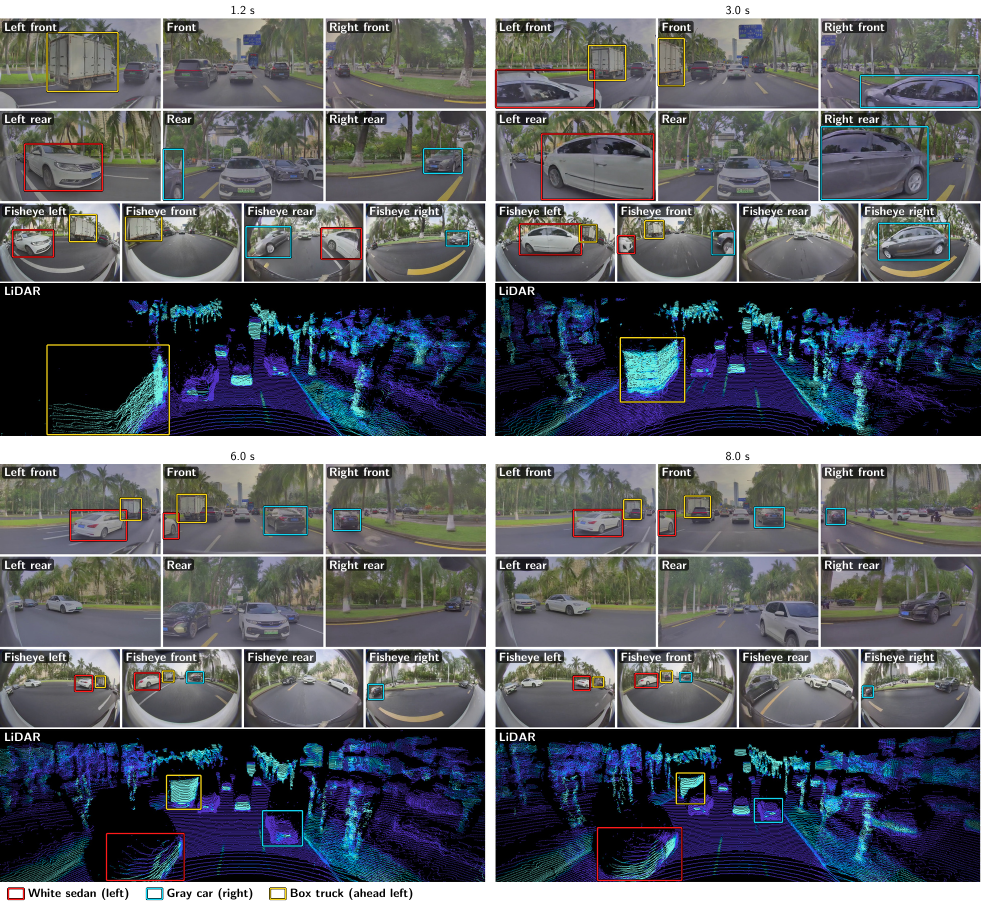}
  \caption{Cross-view and cross-modal consistency over time.
  Panels show layout-conditioned generation for six pinhole cameras, four fisheye cameras, and four LiDARs.
  Matching box colors track the same vehicles across sensors and time: a white sedan overtaking on the left, a gray car overtaking on the right, and the truck ahead.}
  \label{fig:generative-cross-view-consistency}
\end{figure}

\paragraph{Cross-view consistency over time.}
\Cref{fig:generative-cross-view-consistency} shows the results of joint camera and LiDAR generation during a layout-conditioned rollout on an urban street.
Boxes of the same color mark the same object across views and times.
Two vehicles overtake the ego vehicle.
On the left, a white sedan is behind the ego vehicle in the left-rear camera at 1.2\,s, alongside it in the left-rear, left fisheye, and left-front cameras at 3.0\,s, and beside it in the left-front camera at 6.0 and 8.0\,s.
On the right, a gray car approaches in the right-rear camera at 1.2\,s, passes through the right-rear, right fisheye, and right-front cameras at 3.0\,s, and stays ahead in the front camera at 6.0 and 8.0\,s.
Each vehicle keeps its color, shape, and lane as it moves from one view to the next, and it enters the next view as it leaves the previous one.
The box truck ahead on the left stays in the left-front and front cameras and appears in the LiDAR as a dense return cluster at the corresponding position, which moves toward the image center and shrinks as the truck pulls ahead, matching its motion in the left-front camera.

\begin{table}[t]
  \centering
  \caption{Gaussian-conditioned camera generation on evaluation scenes with artifacts, re-inserted assets, or both.
  Gaussian render denotes the input render without generation.
  E2E FT denotes end-to-end fine-tuning (\Cref{eq:generative-end-to-end-fine-tuning-loss}).
  The final two rows compare the causal model with and without the Gram loss during fine-tuning.}
  \label{tab:generative-gaussian-conditioned-results}
  \begin{tabular*}{\linewidth}{@{\extracolsep{\fill}\hspace{\tabcolsep}}lrrrrrr}
    \toprule
    \textbf{Model} & \textbf{Steps} & \textbf{PSNR $\uparrow$} & \textbf{SSIM $\uparrow$} & \textbf{LPIPS $\downarrow$} & \textbf{FID $\downarrow$} & \textbf{FVD $\downarrow$} \\
    \midrule
    Gaussian render & -- & 25.56 & 0.872 & 0.327 & 78.3 & 579.2 \\
    Bidirectional & 35 & 27.04 & 0.852 & 0.288 & 26.3 & 214.9 \\
    Causal & 4 & 26.26 & 0.841 & 0.300 & 26.1 & 226.9 \\
    Causal + E2E FT (w/o Gram) & 4 & \textbf{28.30} & \textbf{0.879} & \textbf{0.248} & 28.0 & 279.1 \\
    Causal + E2E FT (with Gram) & 4 & 27.66 & 0.865 & 0.252 & \textbf{24.7} & \textbf{208.9} \\
    \bottomrule
  \end{tabular*}
\end{table}

\begin{figure}[t]
  \centering
  \includegraphics[width=\linewidth]{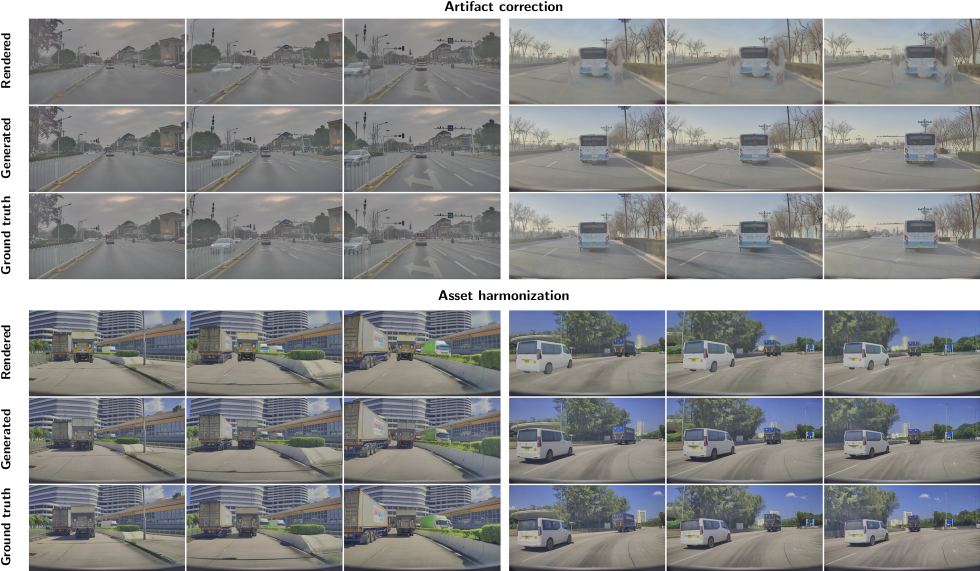}
  \caption{Gaussian-conditioned generation by the final Stage-IV model on front-camera evaluation sequences: the upper two show artifact correction and the lower two show harmonization of inserted assets.
  Each sequence shows three frames, and rows show the Gaussian render, the generated output, and the ground truth.}
  \label{fig:generative-gaussian-conditioned-comparison}
\end{figure}

\paragraph{Gaussian-conditioned generation.}
\Cref{tab:generative-gaussian-conditioned-results} compares the bidirectional and causal models with the Gaussian-render baseline and evaluates end-to-end fine-tuning of the causal model.
We train the bidirectional and causal models at 800 pixels on the long side and fine-tune the causal model end-to-end at 1280 pixels.
The causal model achieves similar FID and FVD to the bidirectional model with fewer denoising steps.
Adding the Gram loss to end-to-end fine-tuning lowers FID from 28.0 to 24.7 and FVD from 279.1 to 208.9, while PSNR decreases from 28.30 to 27.66\,dB.

\Cref{fig:generative-gaussian-conditioned-comparison} shows four front-camera evaluation sequences with three frames each; rows compare Gaussian renders, generated outputs, and ground truth.
The upper two show artifact correction restoring markings, signs, traffic lights, vehicles, and the ghosted bus.
The lower two show harmonization of re-inserted asset renders~\cite{sam3d}, restoring truck and van details and adding the contact shadow.
Corrections remain consistent across frames.

\section{Gaussian Scene Reconstruction}
\label{sec:gaussian-scene-reconstruction}

Gaussian scene reconstruction builds a digital twin of the captured scene from driving logs, providing geometry and appearance grounding for multimodal generation.
At simulation time, \agenticworld{} controls the traffic agents, and renders of the reconstructed scene condition the generator.
In this section, we present a unified representation for camera and LiDAR rendering, controllable 3D assets of agents, and a scalable approach to scene modeling, from individual scenarios to continuous trips within reconstructed regions.

\subsection{Scene representation}
\label{sec:reconstruction-scene-representation}

\paragraph{Gaussian primitives.}
We build on 3D Gaussian Splatting~\citep{kerbl3dgs}, representing the scene with a collection of learnable anisotropic Gaussian primitives.
Each primitive has a center $\boldsymbol{\mu}_i\in\mathbb{R}^3$, an opacity $o_i\in[0,1]$, and a view-dependent color $\mathbf{c}_i$.
A learned decoder maps per-Gaussian features to color, additionally conditioned on a traversal embedding in multi-traversal blocks (\Cref{sec:reconstruction-large-scale}).
Its covariance is $\Sigma_i=R_iS_iS_i^\top R_i^\top$, where $R_i\in SO(3)$ gives orientation and $S_i=\operatorname{diag}(s_{i1},s_{i2},s_{i3})$ contains positive principal-axis scales.
At a location $\mathbf{x}$, its unnormalized spatial response is:
\begin{equation}
  \mathcal{G}_i(\mathbf{x})=\exp\!\Big(-\tfrac12(\mathbf{x}-\boldsymbol{\mu}_i)^\top\Sigma_i^{-1}(\mathbf{x}-\boldsymbol{\mu}_i)\Big).
  \label{eq:reconstruction-gaussian-primitive}
\end{equation}
Rendering combines opacity-weighted Gaussian responses by front-to-back alpha compositing (\Cref{eq:reconstruction-alpha-compositing}).
Camera and LiDAR share this representation, with sensor-specific projection and response evaluation (\Cref{sec:reconstruction-sensor-native-rendering}).

\paragraph{Compositional scene graph.}
Following compositional neural sensor simulation~\citep{neural-scene-graphs,unisim,splatad}, the scene comprises a static background, foreground agent assets, and an ego-vehicle node.
Agent dynamics (\Cref{sec:agent-dynamics}) drive foreground placement, while the simulation runtime maintains the active agents (\Cref{sec:agent-dynamics-model-design,sec:agent-generation}).
The Gaussian set at time $t$ is:
\begin{equation}
  \mathcal{P}(t)=\mathcal{P}_{\mathrm{bg}}
  \cup\Phi_{\mathrm{ego},t}(\mathcal{P}_{\mathrm{ego}})
  \cup\bigcup_{k\in\mathcal{A}(t)}\Phi_{k,t}(\mathcal{P}_k),
  \label{eq:reconstruction-scene-composition}
\end{equation}
where $\mathcal{P}_{\mathrm{bg}}$, $\mathcal{P}_{\mathrm{ego}}$, and $\mathcal{P}_k$ contain background, ego, and agent-$k$ Gaussians, and $\mathcal{A}(t)$ indexes the active agent nodes at time $t$.
The mappings $\Phi_{k,t}$ and $\Phi_{\mathrm{ego},t}$ transform Gaussians from the agent's canonical frame and ego frame, respectively, to world coordinates using their poses at time $t$.
Camera and LiDAR rendering share this compositional representation, with occlusion resolved across nodes during compositing.

\paragraph{Background and ego representation.}
Background Gaussians remain fixed in world coordinates and represent the static surroundings.
Ego Gaussians are defined in the ego frame and transformed by the ego pose, capturing the vehicle structure visible to surround cameras.
The sky is modeled separately by a direction-conditioned far-field model, outside the finite-depth Gaussian set; sky-mask supervision discourages background Gaussians from explaining sky pixels.

\paragraph{Foreground representation.}
Each agent is represented by Gaussians in its own canonical frame.
For a rigid agent $k$, the canonical frame is aligned with its bounding box, and an object-to-world pose $T_k(t)=[R_k(t)\mid\mathbf{t}_k(t)]\in SE(3)$ defines $\Phi_{k,t}$.
The mean and covariance of Gaussian $i$ transform as:
\begin{equation}
  \boldsymbol{\mu}_{ki}^{w}(t)
  =R_k(t)\boldsymbol{\mu}_{ki}^{\mathrm{can}}+\mathbf{t}_k(t),\qquad
  \Sigma_{ki}^{w}(t)
  =R_k(t)\Sigma_{ki}^{\mathrm{can}}R_k(t)^\top.
  \label{eq:reconstruction-agent-rigid-transform}
\end{equation}
For pedestrians, Gaussians are attached to an SMPL body with instance shape $\boldsymbol{\beta}_k$ and time-varying joint pose $\boldsymbol{\theta}_k(t)$ similar to~\cite{omnire}.
Linear blend skinning supplies local articulation, so $\Phi_{k,t}$ includes both body deformation and global placement.
This keeps the asset's shape and appearance separate from its motion through the scene.

Recorded tracks initialize agent motion during reconstruction, with poses refined jointly with geometry and appearance.
Bounding-box assignment and dynamic-mask supervision separate agent content from the background during fitting (\Cref{sec:reconstruction-gaussian-learning}).

\subsection{Sensor-native rendering}
\label{sec:reconstruction-sensor-native-rendering}

Camera images and LiDAR returns supervise the same scene representation through a shared differentiable projection and compositing core, with camera-specific optical models and LiDAR-specific beam sampling.

\subsubsection{Shared rendering core}

\paragraph{Unscented-transform projection.}
We adopt the 3D Gaussian Unscented Transform (3DGUT)~\citep{wu3dgut} to propagate each Gaussian through a nonlinear sensor projection.
For a Gaussian with mean $\boldsymbol{\mu}$, rotation columns $\mathbf{r}_k$, and axis scales $s_k$, the transform constructs $2D+1=7$ sigma points ($D=3$),
\begin{equation}
  \mathbf{x}^{(0)}=\boldsymbol{\mu},\qquad
  \mathbf{x}^{(k,\pm)}=\boldsymbol{\mu}\pm\sqrt{D+\lambda}\,s_k\,\mathbf{r}_k,
  \quad k=1,2,3,
  \label{eq:reconstruction-gaussian-sigma-points}
\end{equation}
where $\lambda=\alpha_{\mathrm{UT}}^2(D+\kappa)-D$. The scaling parameters $\alpha_{\mathrm{UT}}$ and $\kappa$ determine how far the sigma points lie from the mean.
Each point is passed through the calibrated sensor projection, and weighted moments recover the projected distribution.
This construction handles nonlinear optics and spherical projection without a Jacobian linearization or image pre-undistortion.
The projected distribution identifies candidate Gaussians for each sensor sample; their responses are evaluated for compositing.

\paragraph{Compositing.}
For each sensor sample, contributing primitives are ordered front to back.
Their visibility weights and the rendered camera color are:
\begin{equation}
  w_i=\alpha_i\prod_{j<i}(1-\alpha_j),\qquad
  \widehat{X}(\mathbf{p})=\sum_i w_i\,\mathbf{c}_i,
  \label{eq:reconstruction-alpha-compositing}
\end{equation}
where $\mathbf{c}_i$ is evaluated for the current view and traversal appearance condition, and $\alpha_i$ combines primitive opacity with its response along the ray.
The same visibility weights determine the LiDAR range estimates in \Cref{eq:reconstruction-lidar-range-estimates}.
Both modalities differentiate through the shared geometry and agent transforms.

\paragraph{Motion compensation.}
Rolling-shutter cameras and scanning LiDAR acquire samples at different times within a frame.
We account for sensor motion by evaluating the sensor pose at each sample's acquisition time.
Given sensor poses $T_s=[R_s\mid\mathbf{t}_s]$ and $T_e=[R_e\mid\mathbf{t}_e]$ at the interval endpoints, the pose at relative time $\tau\in[0,1]$ is interpolated as:
\begin{equation}
  T(\tau)=\big[\operatorname{Slerp}(R_s,R_e,\tau) \mid (1-\tau)\,\mathbf{t}_s+\tau\,\mathbf{t}_e\big].
  \label{eq:reconstruction-sensor-pose-interpolation}
\end{equation}
Here $T(\tau)$ denotes the world-to-sensor transform used for projection.
Since acquisition time depends on the projected coordinate, each sigma point is projected by solving:
\begin{equation}
  \mathbf{u}=\pi\!\big(T(\tau)\,\mathbf{x}\big),\qquad \tau=\gamma(\mathbf{u}),
  \label{eq:reconstruction-sensor-projection-fixed-point}
\end{equation}
with fixed-point iteration.
The function $\gamma$ follows the sensor's acquisition schedule: image rows for rolling-shutter cameras and recorded timestamps for LiDAR returns.

\subsubsection{LiDAR rendering}

We represent LiDAR measurements as a spherical range image on the calibrated beam grid, capturing the elevation sampling, azimuth resolution, and missing returns that define the resulting point-cloud structure.

\paragraph{Spherical projection.}
Rows of the range image index calibrated beam elevations; columns sample azimuth at the sensor's scan resolution.
Each cell $(r,c)$ represents an ideal beam at elevation $\theta_r$ and azimuth $\phi_c$, with unit direction:
\begin{equation}
  \mathbf{d}(\theta_r,\phi_c)=\big[\cos\phi_c\cos\theta_r,\ \sin\phi_c\cos\theta_r,\ \sin\theta_r\big]^{\top}.
  \label{eq:reconstruction-lidar-beam-direction}
\end{equation}
To project a 3D point onto the range image, we map coordinate $(x,y,z)$ to azimuth $\phi$, elevation $\theta$, and range $r$:
\begin{equation}
  \pi^{S}(x,y,z)=\begin{bmatrix}\phi\\\theta\\r\end{bmatrix}
  =\begin{bmatrix}\operatorname{atan2}(y,x)\\\arcsin(z/r)\\\sqrt{x^2+y^2+z^2}\end{bmatrix}.
  \label{eq:reconstruction-lidar-spherical-projection}
\end{equation}
For each Gaussian, we transform its seven sigma points (\Cref{eq:reconstruction-gaussian-sigma-points}) into the LiDAR coordinate system and apply this mapping.
The projected samples are $\mathbf{y}^{(k)}=\pi^S(T(\tau_k)\,\mathbf{x}^{(k)})$, with $\tau_k$ determined by \Cref{eq:reconstruction-sensor-projection-fixed-point}.
Their weighted mean and covariance in spherical coordinates are:
\begin{equation}
  \boldsymbol{\mu}_{\mathrm{lidar}}=\sum_k w^{(k)}_m\,\mathbf{y}^{(k)},\qquad
  \boldsymbol{\Sigma}_{\mathrm{lidar}}=\sum_k w^{(k)}_c\,\big(\mathbf{y}^{(k)}-\boldsymbol{\mu}_{\mathrm{lidar}}\big)\big(\mathbf{y}^{(k)}-\boldsymbol{\mu}_{\mathrm{lidar}}\big)^{\top},
  \label{eq:reconstruction-lidar-projected-moments}
\end{equation}
where $w_m,w_c$ are the unscented-transform weights.
The angular coordinates $(\phi,\theta)$ index the range image, and rasterization and tiling use calibrated elevation boundaries to account for non-uniform beam spacing.

\paragraph{Return rendering.}
Using $w_i$ from \Cref{eq:reconstruction-alpha-compositing}, we compute accumulated, expected, and median LiDAR ranges:
\begin{equation}
  D_\mathrm{acc}=\sum_i w_i\,d_i,\qquad
  D_\mathrm{exp}=\frac{D_\mathrm{acc}}{\sum_i w_i},\qquad
  D_\mathrm{med}=\min\Big\{d_k:\textstyle\sum_{i=1}^{k}w_i\geq 0.5\Big\},
  \label{eq:reconstruction-lidar-range-estimates}
\end{equation}
where $d_i$ is the range of primitive $i$ along the beam.
During training, we compare the accumulated range $D_\mathrm{acc}$ with recorded LiDAR ranges.
The median range $D_\mathrm{med}$ selects the first primitive whose cumulative opacity reaches $0.5$, falling back to $D_\mathrm{exp}$ if the threshold is not reached.
During simulation, we use $D_\mathrm{exp}$ and switch to $D_\mathrm{med}$ when they differ by more than $0.5\,\mathrm{m}$.
This typically occurs at depth discontinuities, where the weighted average $D_\mathrm{exp}$ falls between foreground and background surfaces, producing noisy LiDAR renderings.
The renderer also predicts return intensity and the probability of no return (ray-drop).

\subsubsection{Camera rendering}

We render images with each camera's calibrated projection, accounting for its field of view and lens distortion.

\paragraph{Camera projection with distortion.}
The projection $\pi$ supports pinhole models with radial, tangential, and thin-prism distortion; fisheye models with radial distortion; and FTheta models.
For a world-to-camera transform $(R,\mathbf{t})$, the sigma points of each Gaussian (\Cref{eq:reconstruction-gaussian-sigma-points}) project to $\mathbf{y}^{(k)}=\pi(R\mathbf{x}^{(k)}+\mathbf{t})$.
Their weighted mean and covariance in the image plane are:
\begin{equation}
  \boldsymbol{\mu}_{\mathrm{img}}=\sum_k w^{(k)}_m\,\mathbf{y}^{(k)},\qquad
  \boldsymbol{\Sigma}_{\mathrm{img}}=\sum_k w^{(k)}_c\,\big(\mathbf{y}^{(k)}-\boldsymbol{\mu}_{\mathrm{img}}\big)\big(\mathbf{y}^{(k)}-\boldsymbol{\mu}_{\mathrm{img}}\big)^{\top}.
  \label{eq:reconstruction-camera-projected-moments}
\end{equation}
These moments define the Gaussian's image footprint for rasterization.
For rolling-shutter cameras, the time-dependent transform in \Cref{eq:reconstruction-sensor-projection-fixed-point} accounts for sensor motion during acquisition.

\paragraph{Image rendering.}
We compute pixel colors by alpha-compositing the view-dependent Gaussian colors using the visibility weights in \Cref{eq:reconstruction-alpha-compositing}.
The ego node supplies visible parts of the vehicle, while the sky model provides the far-field color.

\subsection{Learning Gaussian representation}
\label{sec:reconstruction-gaussian-learning}

We learn the Gaussian representation from camera images and LiDAR returns using the differentiable renderers in \Cref{sec:reconstruction-sensor-native-rendering}.
The Gaussian geometry and appearance are optimized jointly with sensor and agent poses.

\subsubsection{Initialization}

LiDAR points initialize the scene in metric coordinates, supplemented by structure-from-motion points and random samples beyond LiDAR range.
Tracked bounding boxes assign foreground points to agent canonical frames, while the remaining points initialize the background; pedestrian Gaussians are seeded at SMPL vertices.
Initial colors are sampled by projecting the seeds into the nearest image.

\subsubsection{Joint supervision}

The learning objective combines camera and LiDAR reconstruction, as well as structural regularization:
\begin{equation}
  \mathcal{L}=\underbrace{(1-\lambda_\mathrm{SSIM})\,\mathcal{L}_1+\lambda_\mathrm{SSIM}\,\mathcal{L}_\mathrm{SSIM}}_{\text{camera}}
  +\underbrace{\lambda_\mathrm{depth}\,\mathcal{L}_\mathrm{depth}+\lambda_\mathrm{int}\,\mathcal{L}_\mathrm{int}
  +\lambda_\mathrm{drop}\,\mathcal{L}_\mathrm{drop}+\lambda_\mathrm{los}\,\mathcal{L}_\mathrm{los}}_{\text{LiDAR}}
  +\mathcal{R}.
  \label{eq:reconstruction-joint-objective}
\end{equation}
Camera images are supervised with L1 and SSIM losses.
LiDAR supervision uses L1 for accumulated range (\Cref{eq:reconstruction-lidar-range-estimates}), L2 for intensity, and binary cross-entropy for ray-drop over valid reference cells.
The line-of-sight loss $\mathcal{L}_\mathrm{los}$~\cite{unisim,omnire,urf} penalizes opacity in the free space before each measured return.
We filter out the largest range-related errors to improve training robustness.
The regularizer $\mathcal{R}$ includes normal consistency~\citep{huang2dgs}, sky-mask supervision, and the ground and scene-decomposition constraints described below.

\paragraph{Ground constraints.}
At shallow beam angles, small ground-height errors can produce large LiDAR range errors.
We select ground candidates from the lower height quantile of static points and fit a supporting plane by principal-component analysis.
We apply two penalties to ground Gaussians:
\begin{equation}
  \mathcal{L}_\mathrm{flat}=\mathrm{ReLU}\!\Big(\frac{s_z}{s_{xy}}-\rho\Big),\qquad
  \mathcal{L}_\mathrm{cap}=\mathrm{ReLU}\big(s_\mathrm{max}-\tau_s\big),
  \label{eq:reconstruction-ground-regularization}
\end{equation}
where $s_z$ and $s_{xy}$ denote the vertical and horizontal scales of each ground Gaussian, respectively, $s_\mathrm{max}$ is its largest axis length, and $\rho,\tau_s$ are thresholds.
These penalties limit vertical thickening and growth across structural boundaries.
During simulation, the fitted surface supports ground-height queries for object placement.

\paragraph{Scene decomposition.}
For each scene component $b\in\{\mathrm{dyn},\mathrm{bg},\mathrm{ego}\}$, let $\widehat{X}_b$ and $\widehat{M}_b$ denote its image and accumulated opacity in the composed rendering.
Given the GT image $X$ and input image masks $M_\mathrm{dyn},M_\mathrm{ego}$ for agents and ego, we define $M_\mathrm{bg}=(1-M_\mathrm{dyn})(1-M_\mathrm{ego})$.
Writing $\|\cdot\|_{1,M}$ for L1 error within mask $M$, the three losses are:
\begin{equation}
  \begin{aligned}
    \mathcal{L}_\mathrm{dyn}&=\big\|\widehat{X}_\mathrm{dyn}-X\big\|_{1,M_\mathrm{dyn}}+\big\|\widehat{M}_\mathrm{dyn}-\mathcal{S}(M_\mathrm{dyn})\big\|_1,\\
    \mathcal{L}_\mathrm{bg}&=\big\|\widehat{X}_\mathrm{bg}-X\big\|_{1,M_\mathrm{bg}}+\big\|\widehat{M}_\mathrm{bg}-\mathcal{S}(M_\mathrm{bg})\big\|_1,\\
    \mathcal{L}_\mathrm{ego}&=\big\|\widehat{X}_\mathrm{ego}-X\big\|_{1,M_\mathrm{ego}}+\big\|\widehat{M}_\mathrm{ego}-\mathcal{S}(M_\mathrm{ego})\big\|_1.
  \end{aligned}
  \label{eq:reconstruction-scene-decomposition-losses}
\end{equation}
Here $\mathcal{S}$ denotes binarization followed by box-filter smoothing.
The three component losses enter $\mathcal{R}$ with separate loss weights.
For pedestrians, $\mathcal{R}$ also includes second-order temporal smoothness of joint positions, a canonical-offset penalty keeping Gaussian centers within $0.1\,\mathrm{m}$ of SMPL vertices, and total-variation and magnitude penalties on the voxel-based skinning weights.

\subsubsection{Optimization}

We optimize all scene nodes jointly with Adam, refining Gaussian parameters together with sensor and agent poses.
Stochastic MCMC relocation redistributes Gaussians under a fixed budget.
Relocation is grouped by agent to preserve each object's allocation and node membership.
Dynamic and ego mask supervision guides early training and is then disabled to relax constraints near imperfect mask boundaries.
Pedestrian body poses remain fixed during warmup and are subsequently included in joint optimization.

\subsection{Large-scale reconstruction for virtual rides}
\label{sec:reconstruction-large-scale}

For continuous virtual rides beyond individual scenarios, we scale Gaussian reconstruction to larger regions.
We locally align driving logs and partition the observed region into overlapping blocks.
Within each block, repeated traversals contribute to shared geometry, while traversal-specific embeddings account for differences in capture conditions.
The resulting scene assets can be reused along different routes within the modeled region.

\paragraph{Local trajectory refinement.}
Before Gaussian optimization, scan-to-submap registration refines the ego trajectory.
Each LiDAR sweep is de-skewed using per-point timestamps and interpolated sensor poses.
Agent bboxes identify dynamic points for removal, and point-to-plane ICP aligns the remaining static points to a sliding submap from recent frames.
Before registration, we apply voxel downsampling, ground filtering, and range clipping to each point cloud.

\begin{figure}[t]
  \centering
  \includegraphics[width=\linewidth]{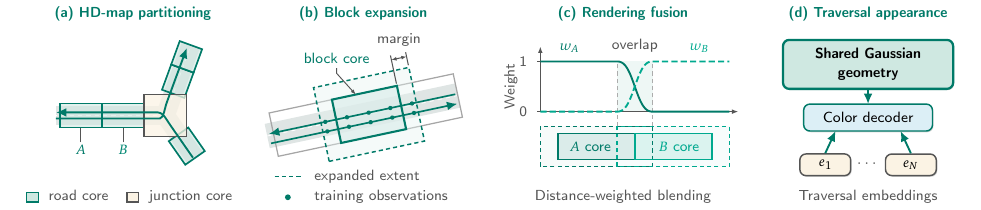}
  \caption{Large-scale scene reconstruction.
  (a) An HD map prior guides partitioning into road and junction block cores.
  Adjacent cores $A$ and $B$ are marked.
  (b) Each block core is expanded by a margin to collect training observations from repeated traversals.
  (c) Schematic blending weights for adjacent blocks $A$ and $B$: full weight in exclusive regions, with distance-based transitions across the overlap region.
  (d) Traversals share Gaussian geometry within a block, while separate appearance embeddings condition color decoding.}
  \label{fig:reconstruction-large-scale}
\end{figure}

\paragraph{Block partitioning and reconstruction.}
Using an HD map prior, we partition the modeled region into blocks that follow the road structure: each junction forms a separate block, while roads between junctions are divided along their centerlines into rectangular blocks of similar length (\Cref{fig:reconstruction-large-scale}a).
We train each block using frames captured within its core and a surrounding margin across all traversals (\Cref{fig:reconstruction-large-scale}b).
Within each traversal, we subsample frames based on distance traveled to reduce redundancy during stops.
Neighboring blocks share frames in their overlapping regions, providing common context near boundaries.
We optimize each block using the objective in \Cref{sec:reconstruction-gaussian-learning}.
The reconstructed blocks can be reused within the modeled region, so virtual rides need not follow previously recorded routes.

\paragraph{Multi-block rendering and fusion.}
To support continuous rendering as the ego vehicle moves between blocks, we blend their outputs across overlapping regions.
For a simulated viewpoint $\mathbf{p}$, we select and load the set $\mathcal{B}(\mathbf{p})$ of blocks whose expanded extents cover $\mathbf{p}$.
A block's exclusive region is the part of its expanded extent not covered by any other block.
When $\mathbf{p}$ lies in this region, we render that block and use its camera and LiDAR outputs directly.
Within an overlap, we render all covering blocks from the same sensor pose and fuse their outputs.
Shared training observations help these independently reconstructed blocks produce consistent renderings.
Let $d_i$ denote the distance from $\mathbf{p}$ to the exclusive region of block $i$.
We assign each block a normalized inverse-square weight:
\begin{equation}
  w_i=\frac{(d_i+\epsilon)^{-2}}{\sum_{j\in\mathcal{B}(\mathbf{p})}(d_j+\epsilon)^{-2}},
  \label{eq:reconstruction-block-fusion-weight}
\end{equation}
where $\epsilon>0$ prevents division by zero.
Weights sum to one and favor blocks whose exclusive regions are closer, shifting contributions as the viewpoint moves through the overlap (\Cref{fig:reconstruction-large-scale}c).
For camera images, fusion is a weighted sum of the block renderings, with $w_i$ applied uniformly to every pixel of block $i$'s image.
For LiDAR returns along the same ray, we average consistent ranges using the block weights and retain a valid return when only one block supplies it.
The runtime updates $\mathcal{B}(\mathbf{p})$ and preloads upcoming blocks as the ego moves.

\begin{table}[t]
  \centering
  \caption{Rendering comparison on the ten nuScenes sequences.
  Best per column is in \textbf{bold}; a dash indicates that rendering is not supported.
  Baseline numbers are taken from SplatAD~\citep{splatad}.}
  \label{tab:reconstruction-nuscenes-benchmark}
  \begin{tabular*}{\linewidth}{@{\extracolsep{\fill}\hspace{\tabcolsep}}lrrrrrr}
    \toprule
    \textbf{Method} & \textbf{PSNR\,$\uparrow$} & \textbf{SSIM\,$\uparrow$} & \textbf{LPIPS\,$\downarrow$} & \textbf{Depth\,$\downarrow$} & \textbf{Intensity\,$\downarrow$} & \textbf{Ray-drop\,$\uparrow$} \\
    \midrule
    UniSim~\citep{unisim}             & 22.66 & 0.743 & 0.414 & 0.060  & 0.063 & --   \\
    NeuRAD~\citep{neurad}             & 26.17 & 0.790 & 0.312 & 0.010  & 0.042 & 93.1 \\
    PVG~\citep{pvg}                   & 24.94 & 0.768 & 0.497 & --    & --   & --   \\
    Street-GS~\citep{streetgaussians} & 25.54 & 0.799 & 0.375 & --    & --   & --   \\
    OmniRe~\citep{omnire}             & 25.50 & 0.798 & 0.386 & --    & --   & --   \\
    SplatAD~\citep{splatad}           & 27.54 & 0.849 & 0.302 & 0.020 & \textbf{0.036} & \textbf{93.8} \\
    \midrule
    Ours      & \textbf{28.36} & \textbf{0.884} & \textbf{0.248} & \textbf{0.005} & \textbf{0.036} & 93.7 \\
    \bottomrule
  \end{tabular*}
\end{table}

\paragraph{Traversal-specific appearance modeling.}
To combine traversals with different lighting, exposure, and weather, we jointly optimize shared block geometry and a separate embedding for each traversal.
The traversal embedding conditions the color decoder to render traversal-specific appearance with shared geometry (\Cref{fig:reconstruction-large-scale}d), merging multiple visits into one reusable asset.

\begin{figure}[t]
  \centering
  \includegraphics[width=\linewidth]{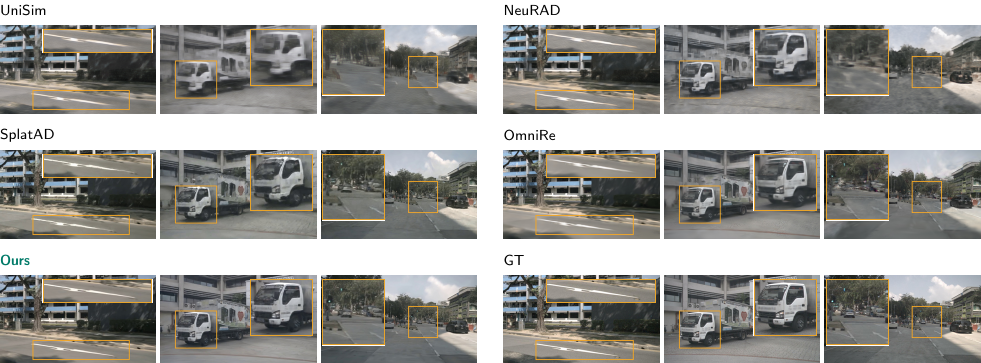}
  \caption{Qualitative comparison of camera renderings on nuScenes.
  Methods are arranged in two columns, with three scenes per method: road markings, vehicle detail, and distant traffic (left to right).
  Insets enlarge the orange-boxed regions; all methods use the same crops and magnification.}
  \label{fig:reconstruction-camera-comparison}
\end{figure}
\begin{figure}[t]
  \centering
  \includegraphics[width=\linewidth]{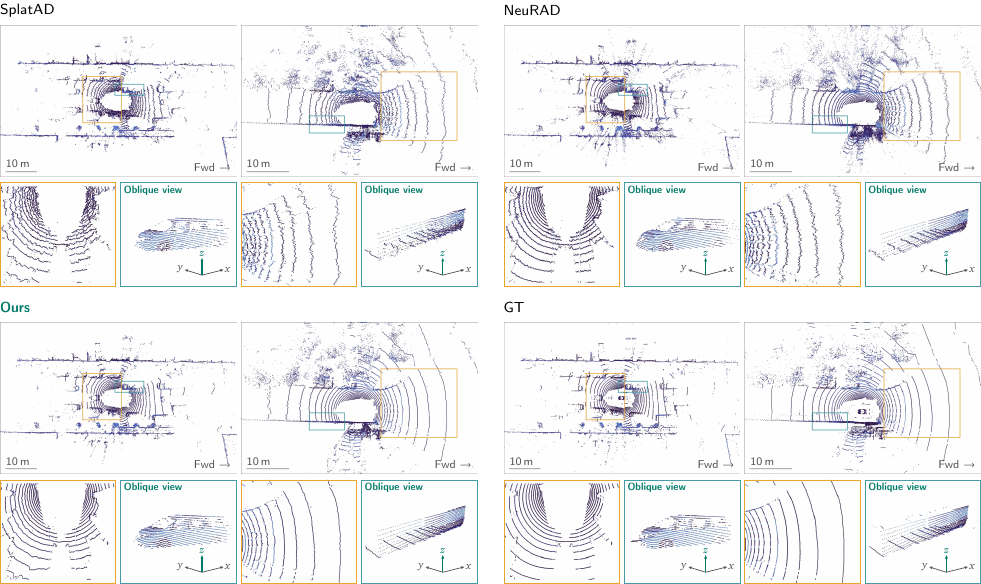}
  \caption{Qualitative comparison of LiDAR rendering on nuScenes.
  Below each top view, the orange-boxed region is enlarged on the left and the teal-boxed region is shown in an oblique view on the right.}
  \label{fig:reconstruction-lidar-comparison}
\end{figure}

\subsection{Model evaluation}
\label{sec:reconstruction-evaluation}

We evaluate reconstruction quality on held-out camera images and LiDAR returns, measuring how accurately the Gaussian scene representation reproduces the appearance and geometry of the real environment.

\paragraph{Evaluation protocol.}
We use the same ten nuScenes sequences as the SplatAD evaluation~\citep{splatad}, with six surround cameras and the top LiDAR.
We hold out every other frame for evaluation and use the remaining frames for training.
Image fidelity is measured by PSNR, SSIM, and LPIPS.
LiDAR fidelity is assessed by the median squared depth error, intensity error, and ray-drop accuracy.
We evaluate our method after $30$k training steps.

\paragraph{Quantitative comparison.}
Among the methods in \Cref{tab:reconstruction-nuscenes-benchmark}, our method achieves the highest PSNR and SSIM and the lowest LPIPS and depth error.
It achieves LiDAR intensity error and ray-drop accuracy comparable to those reported by SplatAD.

\paragraph{Qualitative comparison.}
\Cref{fig:reconstruction-camera-comparison} compares our camera renderings with the baseline methods on the same frames, while \Cref{fig:reconstruction-lidar-comparison} compares LiDAR rendering results.
In camera views, the baselines blur or distort fine structure such as road markings, details above a truck cab, and distant vehicles and traffic lights, while our renderings stay closer to the GT.
In LiDAR comparison, the baselines produce jagged scan lines and scattered returns on the ground, vehicles, and roadside structures, while our renderings are smoother and more faithful to GT.

\section{Applications}
\label{sec:applications}

We apply \methodname{} to closed-loop scenario testing and continuous virtual rides.
The visual results pair generated camera and LiDAR observations with autonomy responses, from yielding to other road users in a local encounter to navigating successive road contexts and traffic interactions along a continuous route.

\subsection{Autonomy Stack Integration}
\label{sec:applications-autonomy-stack-integration}

We integrate \methodname{} with the same autonomy stack deployed on Apollo Go robotaxis.
The stack receives generated camera and LiDAR observations and navigation guidance, and outputs steering, throttle, and brake commands.
A vehicle dynamics model uses these commands to update the world state.
\methodname{} generates the next sensor observations from this state and feeds them back to the stack, closing the loop.

\subsection{Closed-Loop Scenario Testing}
\label{sec:applications-scenario-testing}

\begin{figure}[b]
  \centering
  \includegraphics[width=\linewidth]{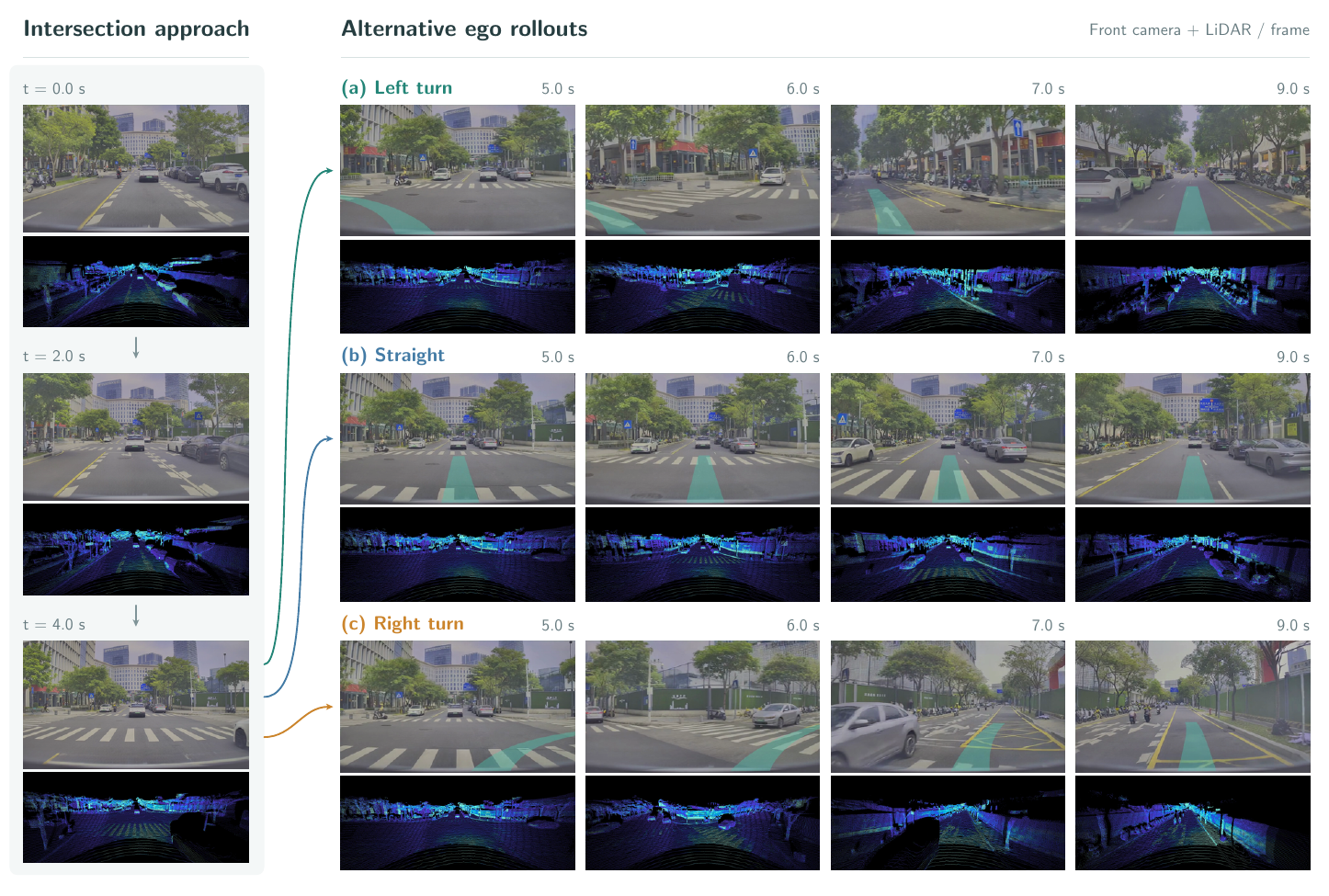}
  \caption{Intersection approach and alternative ego rollouts.
  The left column shows the ego approaching the intersection, using frames from the straight rollout.
  The three rows on the right show (a) left-turn, (b) straight, and (c) right-turn rollouts.
  Each sample pairs generated front-camera (top) and LiDAR (bottom) observations.}
  \label{fig:applications-intersection-maneuvers}
\end{figure}

\paragraph{Alternative driving maneuvers.}
As the ego vehicle approaches an intersection, it may turn left, proceed straight, or turn right.
\Cref{fig:applications-intersection-maneuvers} shows separate rollouts of these three possible futures, with paired front-camera and LiDAR observations showing how the road scene evolves during each maneuver.
These rollouts support scenario testing by showing how different driving choices affect the subsequent road scene and the ego's interactions with surrounding traffic.

\begin{figure}[t]
  \centering
  \includegraphics[width=0.96\linewidth]{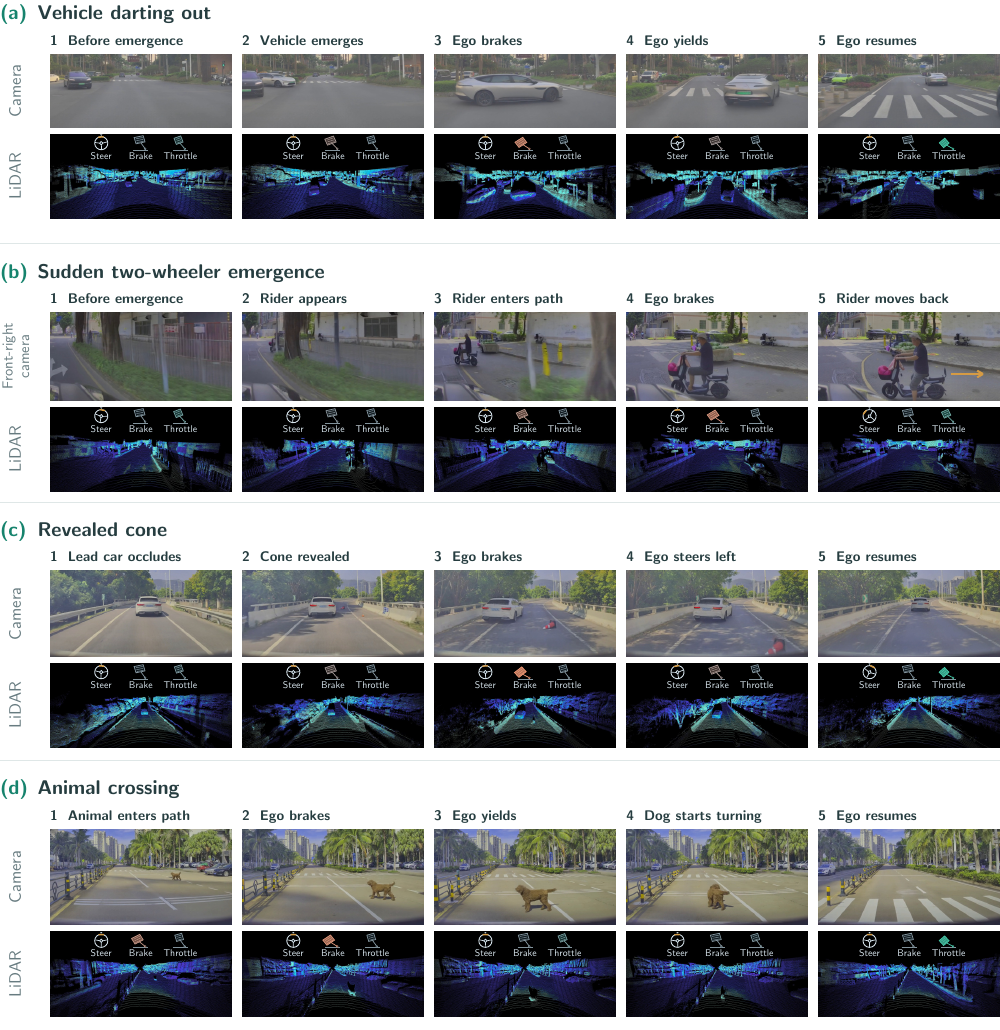}
  \caption{Autonomy responses to safety-critical encounters; read five moments from left to right in each row.
  (a) The ego brakes and yields to an emerging car.
  (b) The ego brakes for an emerging two-wheeler; the rider responds to the approaching ego by backing up.
  (c) The lead car's rear lights turn on as it maneuvers around a fallen cone; the ego brakes and resumes.
  (d) The ego brakes and yields to a crossing dog.
  Paired camera and LiDAR observations with steering, brake, and throttle feedback illustrate closed-loop autonomy responses.}
  \label{fig:applications-safety-critical-encounters}
\end{figure}

\paragraph{Safety-critical scenarios.}
We use closed-loop rollouts to examine braking and yielding in safety-critical scenarios (\Cref{fig:applications-safety-critical-encounters}), including an emerging car or two-wheeler, a fallen cone revealed by a leading vehicle, and a crossing dog.
Overlaid steering, brake, and throttle feedback shows the ego's response; color shading shows braking (orange) and throttle (green) levels.
In the two-wheeler encounter, the ego slows to a near stop while the rider backs up in response to the approaching ego.
In the cone rollout, the leading vehicle's rear lights turn on as it maneuvers around the obstacle.
Our multimodal generation model captures this lighting change together with the vehicle's motion.

\begin{figure}[t]
  \centering
  \includegraphics[width=\linewidth]{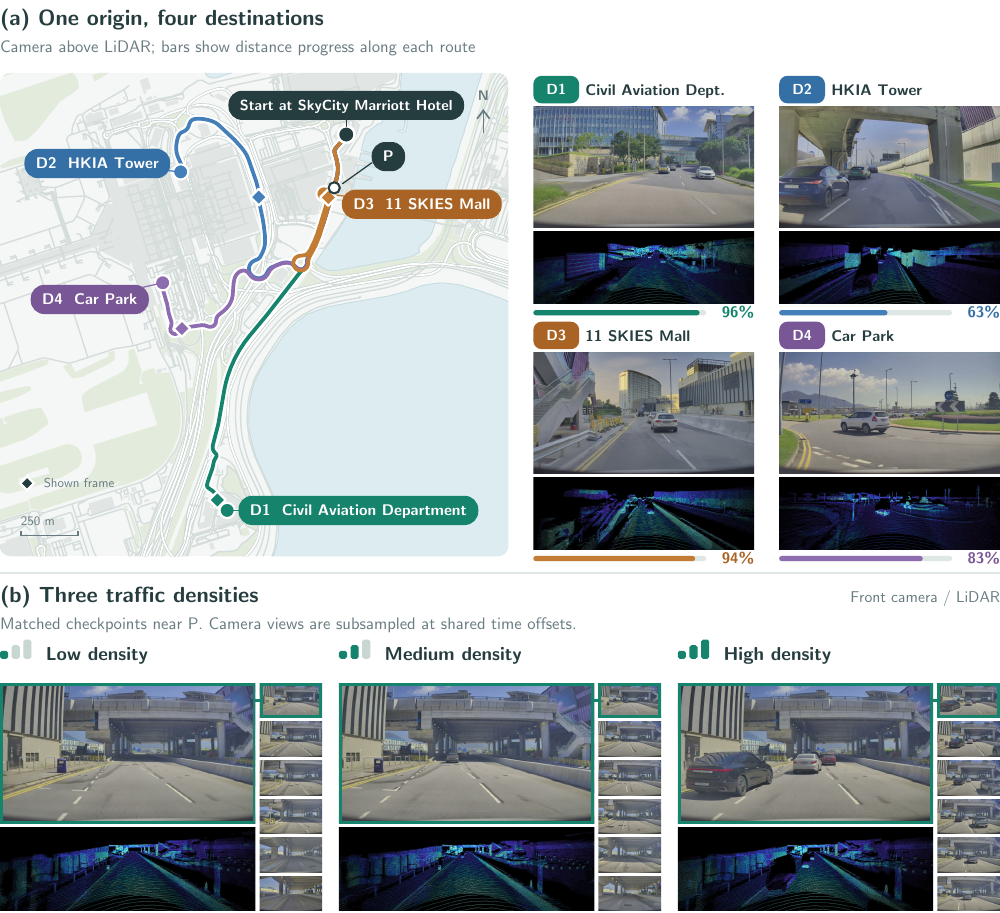}
  \caption{Virtual rides with front-camera and LiDAR generation.
  (a) Four routes share one origin; D1--D4 mark destinations and diamonds locate the displayed frames.
  Bars show distance progress along each route.
  (b) Three closed-loop runs at different densities have longitudinally matched checkpoints near P.
  Each sequence shows six frames at the same elapsed times from its checkpoint.
  The enlarged view is paired with LiDAR.}
  \label{fig:applications-virtual-rides}
\end{figure}

\subsection{Continuous Virtual Rides}
\label{sec:applications-virtual-rides}

Continuous virtual rides extend scenario testing to complete origin-to-destination trips.
Navigation specifies the route, while the autonomy stack chooses local trajectories and control actions from generated camera and LiDAR observations.
Ego and traffic states persist without resets between encounters, so earlier decisions affect later interactions and progress toward the destination.
These trips provide a setting for evaluating sustained driving behavior and autonomy generalization to new cities and operating regions before deployment, enabling reduced reliance on on-road testing during service expansion.
\Cref{fig:applications-virtual-rides}a shows four routes from Hong Kong SkyCity Marriott Hotel to Civil Aviation Department Headquarters (D1), HKIA Tower (D2), 11 SKIES Mall (D3), and Car Park (D4).
Diamonds locate the paired front-camera and LiDAR frames, and bars show the fraction of each route traveled.
\Cref{fig:applications-virtual-rides}b compares low, medium, and high traffic densities at matched longitudinal positions on the same road segment, supporting tests under both rush-hour and off-peak traffic conditions.

\section{Related Work}
\label{sec:related-work}

World models simulate how environments evolve and respond to actions, allowing autonomous systems to explore alternative futures and test decisions beyond recorded scenarios.
In autonomous driving, this requires modeling realistic traffic interactions and generating multimodal sensor observations that remain consistent with the evolving scene.
Below, we review agent-driven world models, learned traffic simulation and scene generation, multimodal generative world models, and reconstruction-based simulation and generative grounding.

\subsection{Agent-Driven World Models}
\label{sec:related-work-agent-driven-world-models}

Recent work explores world models in which agents drive world evolution and generative models synthesize the resulting observations.
Some systems use large language model (LLM) agents to formulate or revise world rules; the resulting programs maintain and update world states, whose structural renderings condition video generation~\citep{code-world-model,coder,matrix-game4}.
Clay-render referencing in Seedance 2.5~\citep{seedance25} provides a related interface for guiding video generation with scene structure, subject motion, and camera placement.
Multi-agent world models jointly condition generation on multiple participants' actions, supporting interaction and consistent observations of a shared world~\citep{solaris,gamma-world}.
Our work shares the principle of agent-driven world evolution.
We learn surrounding traffic agents' responses to the ego and one another from driving data, and use their resulting states to condition multi-view camera and LiDAR generation for sustained closed-loop driving with control over agent states.

\subsection{Learned Traffic Simulation and Scene Generation}
\label{sec:related-work-traffic-simulation}

Learning-based traffic simulation learns how traffic scenes evolve from driving logs.
Several learned simulators predict continuous trajectories or control signals through latent-variable policies, learned state transitions, or hierarchical behavior models~\citep{trafficsim,simnet,mixsim,bits}.
Another line formulates motion generation as autoregressive prediction over discrete vocabularies, exploring trajectory patches, unified scene tokens, and model scaling~\citep{trajeglish,smart,motionlm,behaviorgpt,gpd1,drivegpt,trajtok}.
Our \scenedynamics{} model adopts this tokenized autoregressive formulation for reactive traffic simulation, generating joint agent rollouts conditioned on the evolving scene.

To train traffic agents for sustained interaction, prior work explores closed-loop supervised finetuning, reinforcement learning, adversarial imitation, and self-play~\citep{catk,smart-r1,rift,rlftsim,decompgail,rtr,asymmetric-selfplay}.
Controllability is a separate requirement: simulators should support interventions on intent, interaction style, or scenario outcomes.
Prior work exposes these controls through relations and destinations~\citep{intersim,trafficbots}, diffusion guidance and constraints~\citep{ctgpp,safesim,scenediffuser}, return-conditioned policies~\citep{ctrl-sim}, adversarial trajectory perturbations~\citep{advsim}, or language-derived constraints~\citep{scenario-orchestration}.
Our model combines rollout-based finetuning with goal guidance and behavior sampling, retaining explicit agent states for intervention independently of sensor synthesis.
Runtime checks reject implausible agent responses, and a lightweight planner searches for a valid fallback response.

Scene generation determines which agents and encounters populate the simulated world.
Prior methods generate initial agent states conditioned on a map~\citep{scenegen2021}, or generate both initial states and future trajectories~\citep{trafficgen,unigen}.
Generation can be controlled through spatial constraints~\citep{scenecontrol}, language and scenario descriptors~\citep{lctgen,scenariodiffusion}, retrieved examples~\citep{realgen}, or goal-conditioned denoising~\citep{nexus}.
ScenarioControl connects text- or image-conditioned vectorized scenes to behavior rollouts and camera video synthesis~\citep{scenariocontrol}, while other generators synthesize road networks alongside traffic configurations~\citep{drivescenegen,sledge}.
For evolving traffic populations, SceneDiffuser++, InfGen, and SceneStreamer jointly model motion and agent insertion and removal~\citep{scenediffuserpp,infgen,scenestreamer}.
Scenario Dreamer pairs scene generation with a separate autoregressive behavior model~\citep{scenariodreamer}, and VectorWorld combines streaming map--agent completion and a reactive agent policy~\citep{vectorworld}.
Our \agenticworld{} uses generated sensor observations to select agent insertion regions, including occluded space, avoiding abrupt appearances in previously observed free space.
Conditioned on the road layout and existing traffic, \scenegen{} uses few-step MeanFlow sampling~\citep{meanflow} to populate these regions.
As the ego travels, \scenegen{} adds new agents when appropriate and \scenedynamics{} models their interactions with the ego and one another, while a fallback module handles invalid agent responses, supporting long closed-loop rollouts with evolving traffic.

\subsection{Multimodal Generative World Models}
\label{sec:related-work-multimodal-generation}

Video and world foundation models provide appearance and motion priors~\citep{wan,cosmos-predict25,cosmos3}, which can be adapted to driving through spatial control and domain-specific training~\citep{cosmos-drive-dreams,omnidreams}.
Driving world models support controllable multi-camera generation~\citep{gaia2,xworld,xiaomi-world-model2,omnidreams,drivewm,magicdrive-v2,unimlvg,geniedrive}.
Ego-action-conditioned generation supports flexible control over traffic and appearance through structured conditions or semantic prompts~\citep{gaia2,xworld,causaldrive,geodrive}.
OmniDreams and the Waymo World Model are particularly relevant to our system, integrating controllable sensor generation with closed-loop driving simulation~\citep{omnidreams,waymo-world-model}.
Our simulator explicitly models agent responses to the ego and one another, then generates multimodal observations conditioned on the evolving scene.

Autonomous driving systems rely on multimodal sensor measurements for perception and decision-making, so generative simulation must produce consistent observations across sensors.
Geometric conditioning and cross-view modeling support consistency across camera views~\citep{magicdrive-v2,cogen-drive,gaia2,raynova}.
Multimodal generation further requires different sensing modalities to agree on object positions, geometry, and motion over time.
Prior work studies radar synthesis~\citep{gaia-4,radargen} and LiDAR generation and forecasting~\citep{lidar-diffusion,omnilidar,copilot4d,lidarverse}.
Camera--LiDAR world models jointly generate visual and geometric sensor sequences~\citep{genesis-drive,m4world,waymo-world-model}.
This consistency matters in closed-loop simulation, where sensor discrepancies can alter the autonomy stack's perception and subsequent decisions.
Our work unifies pinhole, fisheye, and LiDAR generation in a single model, addressing consistency across views and modalities as the scene evolves in response to ego actions and interactions among agents.

Closed-loop simulation requires causal prediction, efficient sampling, and robustness to generated histories.
Per-token noise conditioning supports causal sequence diffusion~\citep{diffusion-forcing}; distribution matching and asymmetric distillation accelerate sampling~\citep{dmd,causvid}; and training on self-generated histories addresses exposure bias~\citep{self-forcing}.
Driving applications extend these techniques to camera and LiDAR generation~\citep{omnidreams,causaldrive,lidarverse}.
Longer generation is studied through extended-rollout training~\citep{self-forcing-pp,horizondrive} and inference-time visual memory~\citep{worldtrace}.
We adopt bidirectional training, causal adaptation, and self-forcing distillation for synchronized multimodal generation (\Cref{sec:generative-training}).
We also support optional conditioning on 3D Gaussian reconstructions, which serve as persistent scene memory to help maintain consistent geometry and appearance over long rollouts.

\subsection{Reconstruction-based Simulation and Generative Grounding}
\label{sec:related-work-reconstruction}

Neural reconstruction brings real-world geometry and appearance into simulation.
Neural radiance fields reconstruct driving scenes from sensor logs, with compositional models enabling foreground--background separation and controllable rendering~\citep{urf,neural-scene-graphs,unisim,neurad,emernerf}.
Gaussian splatting provides efficient rendering~\citep{kerbl3dgs} and has been extended to dynamic driving scenes~\citep{streetgaussians,drivinggaussian,hugs,pvg,s3gaussian,desire-gs}.
Recent methods jointly render camera images and LiDAR measurements from a shared Gaussian representation~\citep{splatad,simuli,xsim,uni-gaussians,salf}.
Deformable representations model non-rigid  motion~\citep{pedestrian-simulation,s3,omnire,gart,hugs-human,3dgs-avatar}.
Object reconstruction and generation provide reusable 3D assets for controllable simulation~\citep{neusim,cadsim,asset-harvester,genassets}.
Scalable reconstruction explores learned scene priors and spatial partitioning to reduce reconstruction cost and improve scene coverage~\citep{g3r,flux4d,instant-nurec,block-nerf,mtgs}.
Reconstructed environments support closed-loop policy evaluation and reinforcement learning~\citep{neuroncap,hugsim,rad}.
We adopt 3DGUT's unscented-transform~\citep{wu3dgut} for camera and LiDAR rendering.
Our multi-traversal block reconstruction supports continuous virtual rides within the modeled region.

Reconstructed scenes can exhibit rendering artifacts at novel viewpoints and appearance mismatches after object insertion.
Generative models serve as fixers that correct rendering artifacts at extrapolated viewpoints~\citep{drivedreamer4d,streetcrafter,freesim,recondreamer,drivex,difix3d,artifixer,genre}.
Scene editing modifies object placement and traffic layouts~\citep{neural-assets,scenecrafter,streetcrafter,horizonforge}, while harmonization integrates inserted objects into the scene with consistent appearance and lighting~\citep{diffusionharmonizer,r3d2}.
Inverse rendering and relighting enable controlled changes to scene illumination~\citep{fegr,lightsim,urbanir,diffusionrenderer,unirelight}.
Autoregressive video models further address temporal consistency by conditioning restoration on previously generated frames~\citep{artifixer,omnidreams}.
Our simulator evolves the scene with reactive traffic agents and generates sensor observations conditioned on the layouts or reconstructed scene renders for sustained closed-loop simulation.

\section{Conclusion}
\label{sec:conclusion}

\methodname{} turns recorded driving experience into driveable worlds through interacting traffic agents and joint multimodal generation.
\agenticworld{} creates and completes traffic populations and models how agents pursue their goals and respond to the ego vehicle and one another.
Scene layouts guide sensor generation, with Gaussian reconstructions providing additional geometric and appearance cues where available.
Integrated with an autonomy stack, the system supports safety-critical scenario testing and continuous origin-to-destination virtual rides, in which local driving decisions shape later encounters and progress along the route.
These capabilities extend simulation from isolated cases to sustained driving under varied traffic conditions.
By making driving experience controllable and responsive to actions, \methodname{} provides a foundation for evaluating autonomy beyond recorded outcomes and exploring the use of simulated experience for policy learning.
Building on this foundation, future directions include improving how agents are modeled and developing a reliable evaluation model for driving policies.
\agenticworld{} models agents in a structured space of explicit states and motion tokens, which may limit the detail it can capture, while JEPA highlights limitations of pixel-level prediction~\citep{lecun2022}.
This motivates exploring learned representations between structured states and pixels.
While \methodname{} provides an environment for closed-loop rollouts, a reliable evaluation model for driving policies remains to be developed.
The evaluation model should assess interactions between the ego vehicle and surrounding agents, distinguishing policy errors from unrealistic responses by simulated agents.

\appendix
\section{Contributors and Acknowledgments}
\label{sec:contributors}

We are deeply grateful to the team, whose dedication and perseverance made this work possible.

\paragraph{Contributors}
\begin{tabularx}{\linewidth}{@{}*{4}{Y}@{}}
  Chao Ban & Yangyang Chang & Kai Cheng & Yu Guo \\
  Baicheng Li* & Jixuan Li & Nengjia Li & Shuyuan Li \\
  Shurui Liu & Ruilin Lu & Zhongchang Luo & Linwei Song \\
  Mai Su* & Hao Sun & Rihui Sun & Xiaolin Sun \\
  Yufan Sun & Aoze Wang & Xingshuo Wang & Dong Wu* \\
  Huaxing Yang & Jing Yang & Weifeng Yao & Jin Yu \\
  Jingni Yuan & Qishi Zhang & Baozhu Zhao & Mingze Zhao \\
\end{tabularx}

\vspace{-8pt}

\paragraph{Project Leads}
\begin{tabularx}{\linewidth}{@{}*{4}{Y}@{}}
  Liang Wang & Ze Yang & & \\[0.5\baselineskip]
\end{tabularx}

\vspace{8pt}

The asterisk (*) denotes research interns. Names are listed alphabetically by surname.

\paragraph{Acknowledgments}
We thank all the teams involved in fleet operations, data collection and curation, autonomy algorithm development, infrastructure support, system integration, testing, and more.
Their contributions laid the foundation for \methodname{} and helped carry the project from research and design to a simulation platform for developing and evaluating autonomous driving technologies.
We also thank Baidu for its support throughout the project.

\setcitestyle{numbers}
\bibliographystyle{abbrvnat}
{\linespread{1}\small\bibliography{references}}

\end{document}